%% file: EMNLP-2025-CoT-Risk-main.tex
\pdfoutput=1

\documentclass[11pt]{article}

\usepackage[final]{acl}
\usepackage{times}
\usepackage{latexsym}
\usepackage[T1]{fontenc}
\usepackage[utf8]{inputenc}
\usepackage{microtype}
\usepackage{inconsolata}
\usepackage{graphicx}
\usepackage{amsmath}
\usepackage{amssymb}
\usepackage{bm}
\usepackage{booktabs}
\usepackage{multirow}
\usepackage{makecell}
\usepackage{colortbl}
\usepackage{float}

\title{Chain-of-Thought Prompting Obscures Hallucination Cues in Large Language Models: An Empirical Evaluation}

\author{
    \textbf{Jiahao Cheng}\textsuperscript{1} \quad
    \textbf{Tiancheng Su}\textsuperscript{1} \quad
    \textbf{Jia Yuan}\textsuperscript{1} \quad
    \textbf{Guoxiu He}\textsuperscript{1}\thanks{Corresponding author.} \\
    \textbf{Jiawei Liu}\textsuperscript{2} \quad
    \textbf{Xinqi Tao}\textsuperscript{3} \quad
    \textbf{Jingwen Xie}\textsuperscript{3} \quad
    \textbf{Huaxia Li}\textsuperscript{3} \vspace{5px} \\ 
    \textsuperscript{1} East China Normal University, Shanghai, China \\
    \textsuperscript{2}Wuhan University, Wuhan, China \\
    \textsuperscript{3}Xiaohongshu Inc., Shanghai, China \\
    \texttt{chengjiahao@stu.ecnu.edu.cn, gxhe@fem.ecnu.edu.cn}
}

\begin{document}
    \maketitle
    \begin{abstract}

        Large Language Models (LLMs) often exhibit \textit{hallucinations}, generating factually incorrect or semantically irrelevant content in response to prompts. 
        Chain-of-Thought (CoT) prompting can mitigate hallucinations by encouraging step-by-step reasoning, but its impact on hallucination detection remains underexplored. 
        To bridge this gap, we conduct a systematic empirical evaluation. 
        We begin with a pilot experiment, revealing that CoT reasoning significantly affects the LLM's internal states and token probability distributions. 
        Building on this, we evaluate the impact of various CoT prompting methods on mainstream hallucination detection methods across both instruction-tuned and reasoning-oriented LLMs. 
        Specifically, we examine three key dimensions: changes in hallucination score distributions, variations in detection accuracy, and shifts in detection confidence. 
        Our findings show that while CoT prompting helps reduce hallucination frequency, it also tends to obscure critical signals used for detection, impairing the effectiveness of various detection methods. 
        Our study highlights an overlooked trade-off in the use of reasoning. 
        Code is publicly available at: \url{https://github.com/ECNU-Text-Computing/cot-hallu-detect }.

    \end{abstract}
    
    \input{contents/1.introduction}

\input{contents/3.background}
    \input{contents/4.pilot_exp}
    \input{contents/5.experiments}
    \input{contents/6.results}
    \input{contents/7.conclusion}

    \section*{Limitations}

    Due to computational resource constraints, we are unable to conduct experiments on larger-scale LLMs, which limits the generalizability of our conclusions. Moreover, the limitations of current hallucination detection methods prevented us from evaluating our framework on widely used closed-source LLMs. Future research should incorporate such LLMs to achieve a more comprehensive understanding.

    Regarding evaluation metrics, although the use of Exact Match and ROUGE-L to assess response correctness in QA tasks is subject to debate, we select these metrics because of their widespread adoption, computational efficiency, and effectiveness in measuring surface-level accuracy. Future work may consider more specialized metrics to better capture the nuanced challenges of hallucination detection.

    Our findings show that while CoT prompting significantly improves the reasoning abilities of LLMs, it also introduces complexities that undermine existing hallucination detection mechanisms, posing risks in high-stakes applications. This underscores the urgent need to develop more robust hallucination detection methods tailored to reasoning-augmented LLM. Addressing the impact of CoT prompting on hallucination detection is beyond the scope of this study and is left for future work.
    
    \section*{Ethics Statement}
    
    Our study investigates the impact of CoT reasoning on hallucination detection in LLMs. Importantly, this research does not involve the collection or use of personal data. However, we acknowledge the inherent risks associated with LLMs, including potential biases and misinformation. We are committed to conducting our research responsibly, ensuring transparency in our methodologies, and promoting the ethical use of AI technologies.

    \section*{Acknowledgment}
    This work is supported by the National Natural Science Foundation of China (72204087), the Shanghai Planning Office of Philosophy and Social Science Youth Project (2022ETQ001), the Chenguang Program of Shanghai Education Development Foundation and Shanghai Municipal Education Commission (23CGA28), the Shanghai Pujiang Program (23PJC030), Young Elite Scientists Sponsorship Program by CAST (YESS20240562), and the Fundamental Research Funds for the Central Universities, China. We also appreciate the constructive comments from the anonymous reviewers.
    
    \bibliography{EMNLP-2025-CoT-Risk-ref}
    
    \newpage
    \appendix
    \section*{Appendix}

\input{contents/appendix}

\end{document}

%% file: contents/1.introduction.tex
\section{Introduction}
\label{sec:intro}

Large language models (LLMs) have demonstrated remarkable performance across a range of natural language processing (NLP) tasks \citep{qin2024llmSurvey, yan2024useLlm, song2025siamese}, showing a strong ability to follow instructions and answer questions \citep{zhao2024llmSurvey, hadi2024llmSurvey}.
However, LLMs may also exhibit \textit{hallucinations}. Specifically, they may produce factually incorrect or semantically irrelevant answers to a given prompt without indicating any lack of confidence \citep{ji2023haluSurvey, yang2023llmSurvey, hadi2024llmSurvey}.

To reduce hallucinations in LLMs, extensive research has focused on hallucination detection and mitigation methods \citep{ji2023haluSurvey, huang2024haluSurvey, tonmoy2024haluSurvey}. Existing detection methods are built upon various aspects of LLMs, including output consistency (\textit{e.g.}, evaluating the consistency of responses to multiple identical queries) \citep{wang2023selfconsist, manakul2023selfcheck}, analysis of internal states (\textit{e.g.}, computing confidence metrics) \citep{chuang2024dola, chen2024sharpness, sriramanan2024llmcheck}, and self-evaluation of knowledge (\textit{ e.g.}, querying its certainty) \citep{kumar2024verbConfidence, diao2024active}. Regarding hallucination mitigation, chain-of-thought (CoT) prompting improves response reliability and interpretability by guiding LLMs to generate intermediate reasoning steps before arriving at a final answer \citep{wei2022cot, takeshi2024zeroShot}. In contrast to more complex solutions such as fine-tuning LLMs or improving decoding strategies, CoT prompting is widely adopted due to its simplicity and effectiveness \citep{liu2023cotIsSimpler, yang2024cotIsSimpler}. The performance improvements shown by GPT-4o \cite{hurst2024gpt} and DeepSeek-R1 \cite{guo2025deepseek} through enhanced reasoning have further fueled growing research interest in reasoning-centric LLMs.

\begin{figure*}
  \centering
  \includegraphics[width=\textwidth]{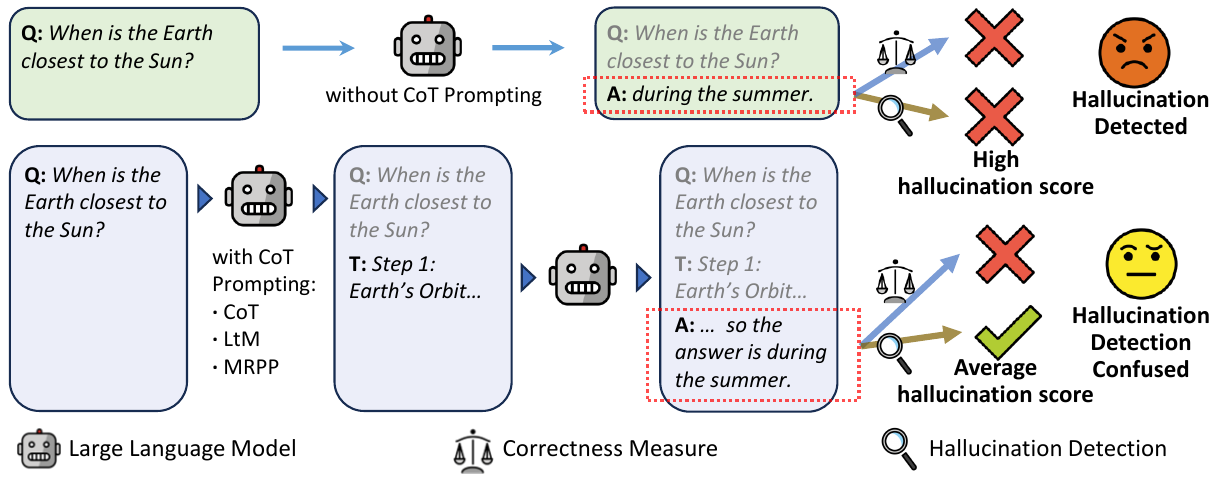}
  \caption{An example of CoT prompting confusing hallucination detection. When the input \textit{When is the Earth closest to the Sun?} is provided to the LLM, it generates a hallucinatory response (\textit{during the summer}) without CoT prompting. In this case, the hallucination detection method assigns a high hallucination score. However, after incorporating CoT prompting, where the LLM introduces reasoning steps (\textit{e.g.}, \textit{Step 1: Earth's Orbit...}), the LLM still produces the same incorrect response, but the detection method assigns only an average hallucination score.}
  \label{fig:example}
\end{figure*}

However, as illustrated in Figure \ref{fig:example}, the use of CoT prompting may diminish the effectiveness of hallucination detection methods. When an LLM is prompted to perform step-by-step reasoning, it semantically amplifies the LLM's internal confidence in its output. As a result, even when the final answer deviates from the ground truth, the LLM tends to produce incorrect tokens with high confidence (\textit{i.e.}, high probability). Consequently, hallucinations induced by CoT prompting often appear more plausible, making them more difficult for detection methods to identify.

Although several counterexamples indicate that reasoning induced by CoT prompting may negatively impact hallucination detection, the generality of this effect has not been established in prior research. To bridge this gap, we conduct a series of empirical evaluations. We start with pilot experiments on three multiple-choice question answering (MCQA) tasks: CommonSenseQA, ARC-Challenge, and MMLU. In these experiments, we observe that CoT prompting significantly alters the probability distribution of the final answer tokens. Building on these findings, we perform a more extensive set of comparative experiments using two representative types of datasets: (1) complex fact-based question answering datasets, including TriviaQA, PopQA, HaluEval, and TruthfulQA; and (2) a summarization dataset requiring deep contextual understanding, named CNN/Daily Mail. We also incorporate four well-established, recently proposed hallucination detection methods to identify hallucinations. Furthermore, we evaluate three mainstream CoT prompting strategies across both instruction-tuned and reasoning-oriented LLMs, including LLaMA-3.1-8B-Instruct, Mistral-7B-Instruct-v0.3, LLaMA-3.1-70B-Instruct, and DeepSeek-R1-Distill-Llama-8B. To comprehensively assess the impact of CoT prompting on hallucination detection, we examine three key dimensions: (1) shifts in the distribution of hallucination detection output scores; (2) changes in detection accuracy; and (3) variations in the confidence levels of the detection methods.

Our experimental results reveal a dual effect of CoT prompting: while it enhances LLM performance, it simultaneously disrupts hallucination detection mechanisms. This disruption arises because CoT prompting systematically alters the distribution of hallucination scores, leading to a counterintuitive trade-off. Particularly, improved task performance is accompanied by reduced hallucination detectability. The extent of this effect varies across detection paradigms: consistency-based methods exhibit greater robustness, whereas internal-state-based and self-evaluation-based methods are more susceptible to reasoning. Notably, advanced detection methods lose their typical advantage over perplexity-based baselines under CoT prompting. These findings suggest that reasoning-enhanced inference reshapes the landscape of hallucination detection, making hallucination signals more subtle and harder to capture.

Our contributions are threefold:
(1) We conduct a series of experiments, including pilot and comprehensive evaluations, to empirically investigate the impact of CoT-induced reasoning on hallucination detection.
(2) We reveal a double-edged effect of CoT prompting: while it enhances LLM performance, it simultaneously obscures common hallucination features.
(3) Our results show that internal-state-based and self-evaluation-based methods exhibit reduced robustness under CoT prompting.

%% file: contents/3.background.tex
\section{Preliminary}

\paragraph{CoT Prompting Methods.}
\label{subsec:prompts}
In this study, we focus on three representative methods: Zero-shot Chain of Thought (\textbf{CoT}, \citet{takeshi2024zeroShot}), Least-to-Most (\textbf{LtM}, \citet{zhou2023ltm}), and Minimum Reasoning Path Prompting (\textbf{MRPP}, \citet{chen2024mrpp}). Further details and formal definitions of these methods are provided in Appendix~\ref{apd:prompts}.

Unless otherwise specified, all CoT prompting methods are implemented in a \textit{zero-shot} configuration. This ensures that the LLM operates without access to task-specific examples, allowing us to isolate the effects of different reasoning strategies. Following prior work, we design prompt templates tailored to each experimental configuration to align with specific task requirements.

\paragraph{Hallucination Detection Methods.}
\label{subsec:halu-detc}
We focus on hallucination detection methods that do not rely on external knowledge bases (\textit{e.g.}, fact-checking \citep{li2024factChecking}) or require additional fine-tuning and specialized training (\textit{e.g.}, ITI \citep{li2023iti} and SAPLMA \citep{azaria2023saplma}). This choice is motivated by the limited generalization ability of such methods across tasks and domains.
Instead, we examine four main categories of hallucination detection methods: \textbf{consistency-based} methods (\textit{e.g.}, SelfCheckGPT-nli \citep{manakul2023selfcheck}); \textbf{internal-state-based} methods (\textit{e.g.}, Perplexity, In-Context Sharpness \citep{chen2024sharpness}, and LLM-Check \citep{sriramanan2024llmcheck}); \textbf{self-evaluation-based} methods (\textit{e.g.}, Verbalized Certainty \citep{kumar2024verbConfidence}); and \textbf{hybrid} methods (\textit{e.g.}, INSIDE \citep{chen2024inside} and SelfCheckGPT-Prompt \citep{manakul2023selfcheck}), which combine features from multiple categories. For comprehensive descriptions and further details of these methods, see Appendix~\ref{apd:halu-detc}.

%% file: contents/4.pilot_exp.tex
\section{Pilot Experiment and Observation}
\label{sec:token-level}

\input{tables/token-level}

% Recent research on multiple-choice question answering (MCQA) tasks \citep{kumar2024verbConfidence, chen2024sharpness, quevedo2024tokenLevel} has revealed a strong correlation between the likelihood of a token being selected by an LLM and the correctness of the final answer.Given that these datasets primarily consist of factual questions, we follow prior work \citep{kumar2024verbConfidence} by using answer incorrectness as a proxy for hallucination labels. Building on these insights, we begin with a pilot experiment on token-level MCQA to examine how CoT-prompted reasoning influences the probability distribution over answer tokens. We pay particular attention to cases where incorrect answers or hallucinations persist despite the use of CoT prompting.  

Recent studies on multiple-choice question answering (MCQA) tasks~\citep{kumar2024verbConfidence, chen2024sharpness, quevedo2024tokenLevel} have identified a strong correlation between the probability assigned by LLMs to answer tokens and the correctness of their final predictions. In these datasets, which are primarily composed of factual questions, we adopt the common practice of treating answer incorrectness as a proxy for hallucination labels~\citep{kumar2024verbConfidence}. Motivated by these insights, we perform a pilot experiment to analyze, at the token level, how reasoning with chain-of-thought (CoT) prompting shapes the distribution of probabilities over possible answer tokens. Our analysis pays special attention to instances where CoT prompting fails to eliminate incorrect (\textit{i.e.}, hallucinated) answers.

\subsection{Pilot Experimental Setting}

\textbf{Datasets}. We use three widely adopted MCQA datasets in our study. \textbf{CommonsenseQA} (validation split, 1,221 samples; \citet{talmor2019commonsenseqa}) is a benchmark designed to evaluate commonsense reasoning; we use the validation split since ground-truth answers are unavailable for its test set. \textbf{AI2 ARC} (ARC-Challenge, test split, 1,172 samples; \citet{boratko2018ai2arc}) is a challenging dataset that focuses on science-related questions requiring advanced reasoning. \textbf{MMLU} (test split, 14,042 samples; \citet{hendrycks2021mmlu}) is a large benchmark that spans a broad range of academic and professional domains, providing a comprehensive assessment of general knowledge and reasoning ability.

\textbf{LLMs}. We conduct the pilot experiments using three open-source LLMs, covering both instruction-tuned and reasoning-oriented variants: \textbf{Llama-3.1-8B-Instruct} \citep{dubey2024llama}, a large-scale model fine-tuned for instruction following; \textbf{Mistral-7B-v0.3-Instruct} \citep{jiang2023mistral}, a compact model trained via instruction tuning; and \textbf{DeepSeek-R1-Distill-Llama-8B} \cite{guo2025deepseek}, a reasoning-oriented model distilled from larger LLMs.

\textbf{Metrics}. We evaluate the LLMs using the following three metrics: \textbf{Accuracy}, which measures overall task performance by assessing the LLM's ability to predict the correct option; \textbf{Entropy}, which quantifies the uncertainty in the LLM's probability distribution over answer options, reflecting its confidence in decision-making; and \textbf{AUROC} (Area Under the Receiver Operating Curve), which treats answer correctness as a proxy hallucination label and evaluates how well the token-level probability distribution distinguishes between hallucinated and non-hallucinated responses.

The implementation details of the pilot experiment are available in Appendix~\ref{apd:details}.

\begin{figure*}[ht]
  \centering
  \includegraphics[width=\textwidth]{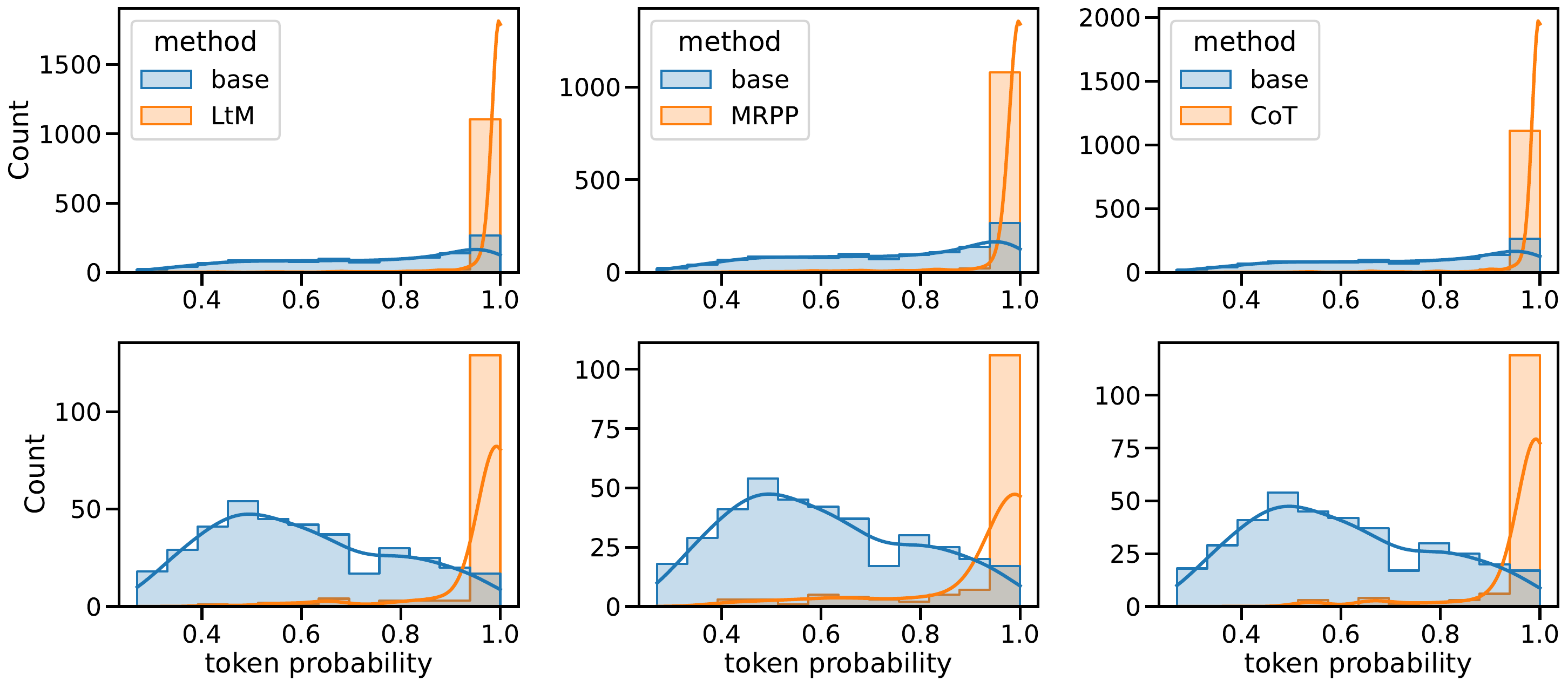}
  \caption{Comparison of token probability distributions for DeepSeek-R1-Distill-Llama-8B on ARC-Challenge. The upper row shows all samples; the lower row shows incorrect predictions after CoT prompting. Kolmogorov-Smirnov test confirms significant differences ($p < 0.01$) between conditions with and without CoT prompting.}
  \label{fig:token-level}
\end{figure*}

\subsection{Observations}

The experimental results are shown in Table~\ref{tab:token-level}. Overall, LLMs prompted with CoT consistently achieve higher accuracy across all datasets. For example, with Llama-3.1-8B-Instruct, CoT improves accuracy (\%) from 90.5, 90.16, and 80.79 to 97.67, 94.96, and 91.57 on the three datasets, respectively. However, the effectiveness of more advanced CoT variants is less consistent. While both LtM and MRPP employ multi-path reasoning, MRPP further incorporates confidence-based validation. As a result, LtM may suffer from ambiguity when faced with multiple plausible reasoning paths, achieving better performance than the base Llama but falling short of standard CoT. In contrast, MRPP outperforms even the basic CoT approach.

In addition, after applying CoT prompting, entropy consistently decreases across all LLMs. This suggests that LLMs become more confident in their predictions by assigning higher probabilities to their selected answers. For instance, Llama's entropy drops from 23.96, 26.33, and 46.94 to 6.63, 13.50, and 21.39, respectively, after applying CoT. Similar reductions are observed with other CoT strategies, reflecting a general improvement in model confidence.

The DeepSeek-R1-Distill-Llama-8B model, which is specifically optimized for reasoning abilities, performs substantially below expectations when reasoning is disabled. Its accuracy on the three datasets drops to 67.41, 59.26, and 54.62, respectively. When CoT is enabled, its performance improves significantly, though it still lags behind Llama in terms of accuracy. Interestingly, under CoT prompting, DeepSeek exhibits lower entropy than Llama across all datasets. This suggests that DeepSeek is more confident in its predictions, even though it produces more incorrect answers.

To better illustrate how confident LLMs are when producing hallucinated answers (\textit{i.e.}, incorrect predictions), we compare their token-level probability distributions before and after applying CoT prompting. This comparison is conducted for both all samples and hallucinated samples. The results are presented in Figure~\ref{fig:token-level}. Before CoT prompting, LLMs show relatively uniform probability distributions across samples. For hallucinated samples, the distributions are clearly left-skewed, indicating that incorrect answers tend to receive lower probabilities. In contrast, after CoT prompting, the probability distributions shift such that LLMs assign similarly high probabilities to both correct and incorrect answers. This increased confidence, especially in wrong predictions, underscores the complex influence of CoT prompting on token-level probabilities, ultimately making hallucinations harder to detect.

Previous studies have shown that token probabilities produced by LLMs do not always reliably indicate the presence of hallucinations \citep{mahaut2024factualConf, lin2022teaching, chen2025hallu}. Instead, more robust signals can be extracted from the internal states of LLMs to better capture hallucination patterns. To explore this, we conduct a more comprehensive set of experiments using recent hallucination detection models. We investigate whether CoT prompting influences the clarity of hallucination-related signals during the reasoning process.

%% file: tables/token-level.tex
\begin{table*}[htbp]
  \centering
  \small
  \caption{Performance comparison of LLMs under CoT prompting on three token-level MCQA tasks. Experimental results show that CoT prompting improves accuracy and reduces entropy, suggesting increased confidence in the LLM's predictions. However, AUROC scores also decline, indicating that token-level probabilities become less reliable as indicators of hallucinations. Detailed results for Mistral-7B-v0.3-Instruct are provided in the Appendix.}
    \begin{tabular}{c|ccc|ccc|ccc}
    \toprule
      & \multicolumn{3}{c|}{ARC-Challenge} & \multicolumn{3}{c|}{CommonSenseQA} & \multicolumn{3}{c}{MMLU} \\
      & Acc & Entropy & AUROC & Acc & Entropy & AUROC & Acc & Entropy & AUROC \\
    \midrule
    \multicolumn{10}{c}{DeepSeek-R1-Distill-Llama-8B} \\
    base & 67.41  & 67.70  & 78.15  & 59.26  & 82.56  & 73.11  & 54.62  & 80.74  & 71.78  \\
    LtM & 87.36  & 5.25  & 69.37  & 73.52  & 6.22  & 69.26  & 76.92  & 11.10  & 68.33  \\
    MRPP & 88.05  & 6.93  & 72.90  & 72.62  & 7.13  & 67.44  & 76.54  & 12.08  & 67.52  \\
    CoT & 87.94  & 4.86  & 70.15  & 72.38  & 5.45  & 68.34  & 77.36  & 10.56  & 64.23  \\
    \midrule
    \multicolumn{10}{c}{Llama-3.1-8B-Instruct} \\
    base & 90.50  & 23.96  & 85.98  & 90.16  & 26.33  & 79.75  & 80.79  & 46.94  & 79.97  \\
    LtM & 93.42  & 18.14  & 78.07  & 89.93  & 27.82  & 76.17  & 85.67  & 36.31  & 78.62  \\
    MRPP & 97.70  & 6.39  & 71.81  & 96.50  & 9.90  & 73.15  & 93.99  & 15.56  & 72.54  \\
    CoT & 97.67  & 6.63  & 77.39  & 94.96  & 13.50  & 74.69  & 91.57  & 21.39  & 76.07  \\
    \bottomrule
    \end{tabular}%
  \label{tab:token-level}%
\end{table*}%

%% file: contents/5.experiments.tex
\section{Main Experiments}
\label{sec:exp}

\begin{figure*}[]
  \centering
  \includegraphics[width=\textwidth]{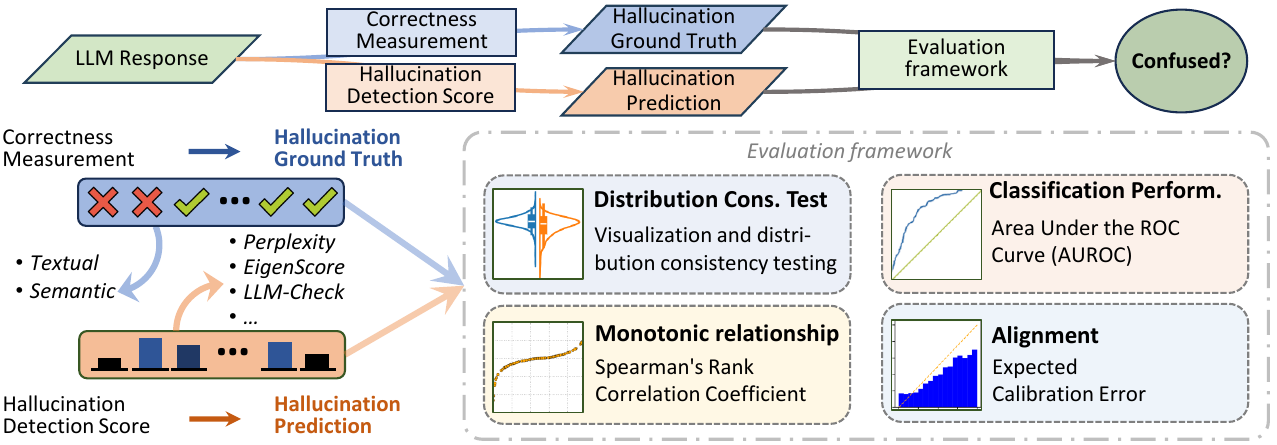}
  \caption{The workflow for evaluating the impact of CoT prompting on hallucination detection. The upper section presents an overview, while the lower section highlights key implementation details. Given the LLM's output, hallucinations are identified by comparing against ground-truth answers. A hallucination detection model then generates a score, which is used to produce predicted labels. Finally, detection performance is evaluated across four dimensions defined in the evaluation framework.}
  \label{fig:methodology}
\end{figure*}

\subsection{Pipeline}
\label{subsec:eval}

This section outlines the workflow used to assess whether CoT prompting interferes with hallucination detection, as illustrated in Figure~\ref{fig:methodology}. We begin by determining whether the LLM-generated response constitutes a hallucination based on the ground-truth response. Next, we use a detection model to compute a hallucination score for each output, which is then used to predict whether the output is a hallucination. Finally, we examine the impact of CoT-generated reasoning on hallucination detection from three perspectives. These include four evaluation metrics.

\textbf{Correctness Measurement}.
For the responses generated by the LLM, we evaluate their alignment with the ground truth in the dataset based on text matching and semantic consistency. This evaluation determines whether a response constitutes a hallucination.

\textbf{Hallucination Detection Score}.
During the response generation process, we apply four mainstream hallucination detection methods to compute the likelihood score of hallucination for each response. Based on these scores, we further predict whether a given response constitutes a hallucination.

\textbf{Evaluation Dimensions}.
Based on the hallucination labels and detection scores, we analyze the behavior of hallucination detection methods along three dimensions: score distribution, detection performance, and method confidence. These dimensions range from surface-level statistical patterns to deeper model-level insights. Specifically: (1) changes in the distribution of hallucination scores are evaluated using a distribution consistency test; (2) detection performance is assessed via AUROC and the monotonic relationship; and (3) method confidence is measured using Expected Calibration Error (ECE). Further details are provided in Appendix~\ref{apd:framework}.

\subsection{Experimental Setup}

\textbf{Datasets.} 
We use four widely adopted QA datasets that require complex reasoning and one summarization dataset. The summarization task demands strong in-context understanding. Specifically, we include TruthfulQA \citep{lin2022truthfulqa}, a dataset explicitly designed for hallucination detection. We use its validation subset, which comprises 817 questions. In addition, TriviaQA \citep{joshi2017triviaqa} is a large-scale reading comprehension dataset. We utilize its \texttt{rc.wikipedia.nocontext} split and validation subset to support our experiments. PopQA \citep{mallen2023popqa} contains questions targeting a long-tail distribution of entities. We use its test subset in a closed-book setting to avoid external knowledge access during evaluation. HaluEval \citep{li2023halueval} evaluates the factual accuracy of LLM-generated content by distinguishing between hallucinatory and accurate outputs. We use its QA subset of non-hallucinatory samples as a benchmark for correctness. For summarization, we employ the CNN/Daily Mail dataset \cite{hermann2015teaching}, where the LLM is required to generate a summary based on a given news article. This task emphasizes faithful summarization, requiring that generated summaries do not deviate from the source content.

\textbf{LLMs.} 
Some hallucination detection methods require access to the internal states of LLMs. Thus, we conduct our experiments using open-source LLMs. To ensure diversity across model families, parameter scales, and training objectives, we select four LLMs, including Llama-3.1-8B-Instruct, Mistral-7B-Instruct-v0.3, Llama-3.1-80B-Instruct, and DeepSeek-R1-Distill-Llama-8B.

Implementation details of the experiments are provided in Appendix~\ref{apd:details}.

%% file: contents/6.results.tex
\section{Results and analysis}
\label{sec:results}

This section presents a comprehensive analysis of the experimental results, focusing on three key aspects: (1) the performance enhancement of LLMs through CoT prompting across diverse tasks and datasets; (2) the impact of CoT prompting on the internal mechanisms of LLMs, as revealed by the distribution shifts in hallucination detection scores; and (3) the consequential effects on the performance and reliability of hallucination detection methods. Our findings demonstrate that while CoT prompting significantly improves output quality, it also alters the models' internal representations and confidence calibration. As a result, this ultimately complicates the hallucination detection process.

We validate our automated metrics by having two annotators manually annotate 150 samples, achieving a Cohen's Kappa of 0.863. The Spearman's $\rho$ between our automated scores and human annotations is 0.690 for Exact Match, 0.656 for ROUGE-L F1, and 0.683 for semantic similarity. These results confirm that our automated evaluations reliably reflect human judgment.

\subsection{Performance Comparison of CoT Methods Across Different Datasets}

As shown in Table~\ref{tab:acc}, the CoT prompting methods significantly enhance the performance of LLMs across various tasks. This is consistent with the results of our pilot experiments. By encouraging LLMs to explicitly generate intermediate reasoning steps, CoT prompting helps activate their latent knowledge and reasoning capabilities, leading to more accurate responses.

Notably, LLM performance rankings vary between the two evaluation approaches, namely textual matching and semantic consistency. This highlights the challenge posed by diverse linguistic expressions in text generation. While evaluation methodology is not the primary focus of this study, we adopt both metrics in subsequent experiments to assess response correctness. This approach helps to mitigate ambiguity, and the resulting correctness serves as the ground truth for hallucination detection.

Furthermore, not all CoT variants consistently improve performance, suggesting that the reasoning process may have nuanced effects on output quality. Interestingly, DeepSeek performs better on summarization tasks, which demand stronger contextual understanding. This observation further supports the notion that incorporating reasoning can enhance LLM performance in such scenarios.

\input{tables/acc}

\subsection{Comparison of Score Distributions for Hallucination Detection} 

To examine the impact of CoT prompting on the internal decision-making processes of LLMs, we employ LLM-Check to compare the distributions of Hidden Score and Attention Score with and without CoT prompts. The Hidden Score reflects the LLM's internal confidence in its own output, while the Attention Score captures the degree of focus allocated to different input segments.

\begin{figure*}[ht]%[ht]
  \centering
  \includegraphics[width=\textwidth]{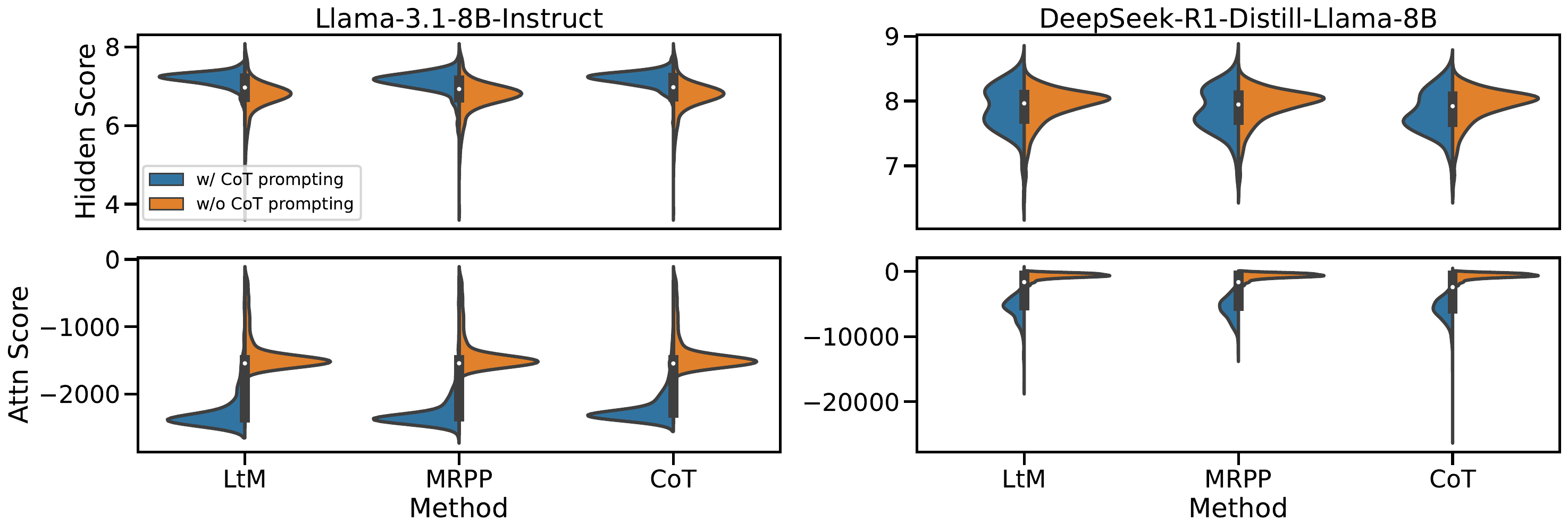}
  \caption{Hidden score and attn score distributions for incorrect predictions. Each plot represents results across different CoT prompting methods (x-axis) and their corresponding scores (y-axis). Significant shifts in score distributions are observed for all hallucination detection methods after incorporating CoT prompting methods, suggesting that CoT prompting methods substantially impact hallucination detection scores: even when hallucinations exist, the scores of hallucination detection methods are biased towards \textit{not hallucination}.}
  \label{fig:dist}
\end{figure*}

We use the Kolmogorov-Smirnov (K-S) test to evaluate whether these score distributions differ significantly between conditions with and without CoT prompting. As shown in Figure~\ref{fig:dist}, both scores exhibit statistically significant shifts ($p < 0.01$), indicating that CoT prompting induces substantial changes in the LLM's internal representations.

For instruction-tuned LLMs such as LLaMA, the scores are relatively dispersed in the absence of CoT prompting, suggesting that hallucination detection methods can effectively differentiate between hallucinated and faithful responses. However, with CoT prompting, the scores become tightly clustered. This implies that the detector may collapse to predicting a single class, thereby losing its discriminative capability.

In contrast, reasoning-oriented LLMs such as DeepSeek exhibit concentrated score distributions even without CoT, suggesting that internal signals are less indicative of hallucination. CoT prompting still significantly alters the distribution, suggesting that explicit reasoning substantially reshapes the LLM's internal state, even when the LLM is already strong in reasoning tasks.

Overall, these findings suggest that CoT prompting not only enhances output quality but also modulates the LLM's internal confidence and attention.

\begin{table*}
  \centering
  \small
  %\caption{Number of experimental configurations (up to 12 per cell: 8 detection methods $\times$ 3 CoT variants) that show a decline in AUROC score for hallucination detection after applying CoT prompting, relative to the base scenario. Configurations are grouped by LLM and dataset.}
  \caption{Number of experimental configurations in which the AUROC for hallucination detection decreases after applying CoT prompting, relative to the base scenario. Each cell can contain up to 12 configurations, corresponding to 3 CoT variants $\times$ 4 LLMs.}
    \begin{tabular}{l|cc|cc|cc|cc}
    \toprule
      & \multicolumn{2}{c|}{HaluEval} & \multicolumn{2}{c|}{PopQA} & \multicolumn{2}{c|}{TriviaQA} & \multicolumn{2}{c}{CNN/Daily Mail} \\
      & Textual & Semantic &Textual & Semantic & Textual & Semantic & Textual & Semantic \\
    \midrule
    PPL &               9 & 6 & 12 & 6 & 10 & 3 & 10 & 8 \\
    Sharpness &         9 & 6 & 12 & 6 & 10 & 4 & 10 & 11 \\
    Eigen Score &       3 & 8 & 4 & 7 & 9 & 6 & 4 & 7 \\
    SelfCk-Prompt &     10 & 7 & 12 & 8 & 9 & 7 & 9 & 6 \\
    SelfCk-NLI &        9 & 6 & 6 & 2 & 12 & 9 & 8 & 4 \\
    Hidden Score &      8 & 9 & 4 & 6 & 8 & 5 & 6 & 7 \\
    Verbalized Certainty & 9 & 7 & 6 & 9 & 11 & 8 & 10 & 7 \\
    Attn Score &        5 & 4 & 0 & 7 & 1 & 5 & 12 & 7 \\
    \bottomrule
    \end{tabular}%
  \label{tab:auroc-all}%
\end{table*}%

\subsection{Performance Comparison of Hallucination Detection Methods} 

% We conduct a comprehensive evaluation of how CoT prompting affects the performance of hallucination detection methods. Our experiments span three CoT prompting methods, four benchmark datasets, four LLMs, two hallucination annotation protocols (\textit{i.e.}, methods for establishing ground truth), and eight representative detection approaches, yielding a total of 768 experimental configurations. 

% Table~\ref{tab:auroc-all} summarizes the proportion of settings in which the AUROC score declined after applying CoT prompting. Overall, more than half of the setups (465 out of 768) exhibit performance degradation, suggesting that CoT prompting exerts a broad negative impact on hallucination detection. 

We conduct a comprehensive evaluation of how CoT prompting affects the performance of hallucination detection methods. Our experiments span four benchmark datasets, four LLMs, two hallucination annotation protocols (\textit{i.e.}, Textual or Semantic methods for establishing ground truth), and eight representative detection approaches. For each combination, we evaluate performance under four prompting scenarios: one base scenario (without CoT) and three distinct CoT prompting methods (CoT, LtM, and MRPP). This results in a total of 768 experimental configurations (4 benchmark datasets $\times$ 4 LLMs $\times$ 2 hallucination protocols $\times$ 8 detection methods $\times$ 3 CoT prompting methods). For each configuration, the baseline performance without CoT prompting serves as the point of comparison.

Table~\ref{tab:auroc-all} summarizes, grouped by LLM and dataset, the number of experimental configurations in which a detection method's AUROC score declined under a CoT prompting scenario compared to the base scenario. Each cell in the table has a maximum value of 12, representing the 4 models evaluated under the 3 CoT methods. Overall, more than half of the configurations (465 out of 768) exhibit performance degradation, suggesting that CoT prompting exerts a broad and substantial negative impact on hallucination detection.

Notably, TruthfulQA is treated separately due to its distinct evaluation protocol. To further assess the generality of our findings, we also evaluate a model-training-based method, MIND \citep{su2024mind}. We observe that it likewise suffers degraded hallucination detection performance under CoT prompting (see Appendix~\ref{apd:results} for details).

Our findings reveal that CoT prompting generally impairs the effectiveness of hallucination detection, with the degree of impact varying by method. Self-evaluation-based approaches (\textit{e.g.}, SelfCheckGPT-Prompt and Verbalized Certainty) are most adversely affected. These methods rely heavily on confidence signals present in the LLM's output. However, these signals are easily distorted by CoT-style reasoning. As a result of CoT prompting, LLMs often generate additional justificatory content, which can obscure factual inaccuracies. For instance, in response to the TriviaQA question \textit{What was Walter Matthau's first movie?}, an LLM under LtM prompting may include phrases like \textit{Based on reliable sources such as IMDb and Wikipedia...}. Although the answer is incorrect, the Verbalized Certainty score may increase from 0.2 to 1.0, illustrating how CoT prompting can mask hallucinations.

By contrast, consistency-based methods (\textit{e.g.}, EigenScore and SelfCheckGPT-NLI) show greater robustness to CoT prompting. These methods leverage multi-sample reasoning to identify inconsistencies, which remain informative even in the presence of CoT prompts. In the above example, consistent sampling occasionally yields the correct answer \textit{The Kentuckian} (EigenScore: 3/15; SelfCheckGPT-NLI: 1/20), albeit infrequently. Although these approaches are often criticized for their high computational overhead, they demonstrate strong resilience and reliability in scenarios that require extensive reasoning.

Another notable observation is that, although advanced hallucination detection methods typically outperform simple perplexity-based baselines under standard conditions, this advantage diminishes when CoT prompting is applied. Without CoT, perplexity-based methods consistently underperform, aligning with prior studies \citep{chen2024inside, sriramanan2024llmcheck, chen2025hallu}. However, when CoT prompts are introduced, many detection methods fail to surpass the perplexity baseline, and some perform worse. This finding underscores a critical limitation of current detection approaches: they are not designed to handle the structural and discourse-level shifts induced by explicit reasoning. As a result, their effectiveness at detecting hallucinations is diminished.

\textbf{TruthfulQA}. As shown in Table~\ref{tab:tqa-roc}, CoT prompting has inconsistent effects on enhancing both the truthfulness and informativeness of responses in the TruthfulQA task across different scenarios. Among the 144 experimental settings, 81 exhibit performance degradation (see Appendix~\ref{apd:results} for details). Meanwhile, the negative impact of CoT prompting on hallucination detection methods varies significantly among different LLMs. Specifically, Llama exhibits a more pronounced drop in informativeness, Mistral experiences a greater decline in truthfulness, and DeepSeek shows a decline in both dimensions. This variation likely reflects differences in response generation tendencies among the various LLMs. For example, following CoT prompting, Llama tends to emphasize the reasoning process and frequently includes redundant phrases, such as \textit{Based on the above steps...}, in its answers. Although such content highlights the reasoning chain, it may obscure factual inaccuracies or introduce misleading elements. This can interfere with the effectiveness of hallucination detection methods. To mitigate potential biases introduced by the LLaMA-based judge model, we additionally fine-tuned the evaluation model using Qwen. The evaluation results based on the Qwen judge model (see Appendix~\ref{apd:results} for details) are consistent with our previous findings, confirming the robustness of our conclusions.

\input{tables/tqa-roc}

\textbf{Monotonic Relationship.}
The trends observed in monotonic relationships are consistent with those in classification performance. Comprehensive experimental results are provided in Appendix~\ref{apd:results}.

\subsection{Confidence in Hallucination Detection Methods}

We further employ Expected Calibration Error (ECE) to evaluate the alignment between the hallucination probabilities predicted by the detection methods and the actual correctness of the hallucination detection. This allows us to assess the calibration of model confidence. The results indicate that CoT prompting leads to an increase in ECE across most experimental settings, suggesting that the confidence of detection methods becomes less reliable. Detailed results are provided in Appendix~\ref{apd:results}.

%% file: tables/acc.tex
\begin{table}[htb]
  \centering
  \small
  \caption{Results of CoT prompting influencing the performance of LLMs across various datasets. The table presents results for Llama-3.1-8B-Instruct and DeepSeek-R1-Distill-Llama-8B on HaluEval, PopQA, TriviaQA, and CNN/Daily Mail (C./D.M.). The results indicate that CoT prompting substantially improves LLM performance on these tasks. Please refer to the appendix for additional tasks and LLMs.}
    \begin{tabular}{ccccc}
    \toprule
      & HaluEval & PopQA & TriviaQA & C./D. M. \\
    \midrule
    \multicolumn{5}{c}{Llama-3.1-8B-Instruct, Textual} \\
    base & 30.26 & 30.31 & 79.41 & 22.13  \\
    CoT & 36.59 & 35.05 & 81.60 & 23.33  \\
    LtM & 36.98 & 33.57 & 80.02 & 22.72  \\
    MRPP & 36.04 & 33.00 & 80.46 & 22.83  \\
    \midrule
    \multicolumn{5}{c}{Llama-3.1-8B-Instruct, Semantic} \\
    base & 39.89 & 48.58 & 85.61 & 91.80  \\
    CoT & 77.37 & 65.30 & 85.17 & 91.07  \\
    LtM & 38.86 & 46.79 & 85.63 & 90.79  \\
    MRPP & 78.10 & 68.03 & 85.60 & 90.82  \\
    \midrule
    \multicolumn{5}{c}{DeepSeek-R1-Distill-Llama-8B, Textual} \\
    base & 19.19  & 16.09  & 61.78  & 20.04  \\
    CoT & 29.21  & 22.09  & 63.16  & 22.54  \\
    LtM & 30.15  & 19.01  & 64.10  & 21.39  \\
    MRPP & 29.21  & 18.79  & 64.34  & 21.68  \\
    \midrule
    \multicolumn{5}{c}{DeepSeek-R1-Distill-Llama-8B, Semantic} \\
    base & 62.30  & 55.47  & 72.28  & 91.26  \\
    CoT & 78.53  & 63.15  & 80.90  & 92.35  \\
    LtM & 79.57  & 64.10  & 83.34  & 91.88  \\
    MRPP & 79.35  & 64.34  & 83.02  & 91.36  \\
    \bottomrule
    \end{tabular}%
  \label{tab:acc}%
\end{table}%

%% file: tables/tqa-roc.tex
\begin{table}[htb]
    \centering
    \caption{Number of experiments of TruthfulQA in which the AUROC value decreases. For each experiment, AUROC is calculated by comparing the predicted hallucination detection scores with either informativeness (I.) or truthfulness (T.); each cell shows the number of experiments (three using CoT prompting and one without) where AUROC decreases relative to a baseline. The \textbf{first two} columns are results for \textbf{Llama}-3.1-8B-Instruct, the \textbf{middle two} for \textbf{Mistral}-7B-Instruct-v0.3, and the \textbf{last two} for \textbf{DeepSeek}-R1-Distill-Llama-8B. } % 
    \begin{tabular}{lcccccc}
    \toprule
      & \multicolumn{1}{l}{I.} & \multicolumn{1}{l}{T.} & \multicolumn{1}{l}{I.} & \multicolumn{1}{l}{T.} & \multicolumn{1}{l}{I.} & \multicolumn{1}{l}{T.} \\
    \midrule
    Perplexity & \textbf{3} & 0 & 0 & \textbf{2} & \textbf{2} & \textbf{3} \\
    Sharpness & \textbf{3} & 0 & 0 & \textbf{2}  & \textbf{2} & \textbf{3} \\
    Eigen Score & \textbf{2} & \textbf{3} & \textbf{1} & \textbf{1}  & 0 & \textbf{3} \\
    SelfCk-Prompt & \textbf{1} & \textbf{3} & \textbf{1} & 0  & \textbf{2} & \textbf{3} \\
    SelfCk-NLI & \textbf{1} & \textbf{2} & 0 & \textbf{3}  & \textbf{2} & \textbf{3} \\
    Verb. Certainty & \textbf{2} & \textbf{2} & 0 & \textbf{2}  & 0 & 0 \\
    Hidden Score & \textbf{2} & 0 & \textbf{3} & \textbf{3}  & \textbf{2} & \textbf{2} \\
    Attn Score & 0 & \textbf{2} & \textbf{3} & \textbf{3}  & \textbf{1} & \textbf{3} \\
    \bottomrule
    \end{tabular}%
    \label{tab:tqa-roc}%
    
\end{table}%

%% file: contents/7.conclusion.tex
\section{Discussion and Conclusion}

In this study, we systematically investigate the impact of CoT prompting on hallucination detection in LLMs. We evaluate several representative detection methods and find that, although CoT prompting improves text-level and semantic accuracy across most datasets, it simultaneously undermines hallucination detection performance. More specifically, CoT prompting results in more concentrated detection score distributions, lower classification accuracy, weaker alignment with ground truth, and poorer confidence calibration. The sensitivity of this impact varies according to the detection method: self-evaluation-based approaches are the most affected, whereas consistency-based methods show greater robustness. We further observe that CoT prompting diminishes and occasionally eliminates the performance advantage of advanced detectors over a simple perplexity baseline.

These findings suggest that CoT prompting can obscure hallucination cues and compromise the effectiveness of existing detection systems, highlighting the need to develop more robust approaches specifically tailored to reasoning-oriented LLMs.

%% file: contents/appendix.tex
\section{Theoretical and Practical Implications}

The theoretical significance of this study lies in its exploration of the relationship between CoT prompting and hallucination detection. 
Our research demonstrates that while CoT prompting methods enhance the interpret-ability and reasoning capabilities of LLMs, they simultaneously confuse the task of detecting hallucinations in these LLMs. 
This highlights a gap in current research: hallucination detection methods must account for the effects introduced by CoT prompting. 
From a practical perspective, our work underscores the importance of considering a broader range of factors in hallucination detection. Specifically, we show that detection methods should be capable of handling the subtle differences introduced by CoT prompting. 
This is crucial for ensuring the safe and reliable deployment of LLMs in real-world applications, especially those requiring high levels of factual accuracy, such as healthcare, legal, and scientific fields. 

\section{Additional Related Work}

\subsection{Large Language Models and Hallucinations}

Large language models demonstrate remarkable generalization capabilities, leading to their widespread adoption across a variety of tasks \citep{zhao2024llmSurvey, qin2024llmSurvey, yan2024useLlm, huang2025textclustering}.
They exhibit strong performance, especially in question-answering tasks \citep{zhao2024llmSurvey,yang2023llmSurvey, hadi2024llmSurvey}.
However, despite their effectiveness, LLMs are not infallible, and a significant challenge that persists is the phenomenon of hallucinations \citep{ji2023haluSurvey, hadi2024llmSurvey, wang2025silentLlms}.

Hallucinations in LLMs typically manifest when the model generates fabricated or inaccurate information in an attempt to answer a question \citep{yang2024alignment, tonmoy2024haluSurvey}.
These hallucinations can be classified into three categories: Input-Conflicting, Context-Conflicting, and Fact-Conflicting hallucinations \citep{zhang2023song}. 
Alternatively, they are often categorized into two broad types: factuality hallucination and faithfulness hallucination \citep{huang2024haluSurvey}. 
Factuality hallucinations arise when the LLM's output conflicts with established world knowledge.
Specifically, these hallucinations arise when the input prompt lacks relevant factual information \citep{he2025selectiveContextual}. 
In such cases, the LLM relies on its internal parameters and memory \citep{mallen2023popqa}, resulting in outputs that contradict factual reality.

In the context of QA tasks, particularly closed-book QA, the issue of Fact-Conflicting or factuality hallucinations is especially pronounced \citep{jin2024tugWar, zhang2023song}.
Several benchmarks are designed to assess these types of hallucinations. Examples include TruthfulQA \citep{lin2022truthfulqa} and HaluEval \citep{li2023halueval}, which explicitly focus on hallucinations. 
Other QA datasets, such as TriviaQA \citep{joshi2017triviaqa} and PopQA \citep{mallen2023popqa}, were not originally designed for hallucination detection but are still widely used to evaluate factual accuracy \citep{chen2024sharpness, chen2024inside, chen2024knowledgeBoundary}.

A variety of methods are proposed to mitigate factuality hallucinations \citep{tonmoy2024haluSurvey}, including prompt engineering, the integration of external knowledge, changes to decoding strategies, and model training or fine-tuning approaches.
However, for users interacting directly with LLMs, prompt-based methods, such as Chain-of-Thought \citep{wei2022cot}, provide a more accessible and widely adopted solution.
The core idea behind CoT is to guide the LLM in generating reasoning steps before providing a final answer, thereby enhancing response accuracy and reducing hallucinations.

\subsection{CoT prompting}

Chain-of-Thought (CoT) prompting is a technique designed to enhance the performance of LLMs by guiding them to generate intermediate reasoning steps prior to producing a final answer \citep{wei2022cot}.
Rather than generating a direct one-step response, CoT prompts the LLM to engage in a structured reasoning process, thereby reducing errors and hallucinations that arise from premature conclusions.
By leveraging this step-by-step reasoning approach, CoT demonstrates significant improvements in both accuracy and reliability \citep{yao2025metacognitive}.

To reduce the manual effort required to provide examples and to better leverage LLMs' prior knowledge, Self-generated Few-shot CoT is introduced \citep{yasunaga2024analogical}. 
This approach allows LLMs to autonomously construct relevant examples and knowledge, enabling adaptive reasoning. 
Remarkably, studies demonstrate that performance improvements associated with reasoning can occur even without explicitly providing examples \citep{takeshi2024zeroShot}.
By simply appending a phrase such as ``let's think step by step'' to the prompt, a technique known as Zero-shot CoT, LLMs are prompted to generate reasoning steps.
This appended phrase (referred to as a ``magic phrase'' \citep{stechly2024thoughtlessness}) is identified as a key trigger for enabling reasoning without explicit demonstrations.

Regardless of whether few-shot examples, self-generated examples, or zero-shot techniques are utilized, the primary objective of these methods is to prompt reasoning prior to delivering a final answer. 
This foundational idea predates the rise of LLMs \citep{nye2022beforeCot, ling2017beforeCot}. 
Beyond Zero-shot CoT, numerous variants emerge to address specific limitations.
For example, Tree-of-Thought (ToT) \citep{yao2023tot} and Graph-of-Thought (GoT) \citep{yao2024got} propose more sophisticated reasoning structures that facilitate backtracking and synergistic outcomes, thereby addressing the linear constraints of CoT reasoning.
Other approaches, such as Least-to-Most Prompting (LtM) \citep{zhou2023ltm} and Minimum Reasoning Path Prompting \citep{chen2024mrpp}, aim to decompose complex problems into simpler sub-problems or ensure that reasoning paths remain as concise and clear as possible.

Collectively, these methods share a common emphasis on generating reasoning steps prior to producing a final answer.
For clarity, we refer to them as \textbf{CoT prompting}. 
They demonstrate broad utility, consistently enhancing LLM performance across various tasks and domains \citep{guo2025useCot, kim2023useCot, nguyen2025useCot}.
Substantial research examines both the empirical and theoretical effectiveness of CoT prompting \citep{feng2023cotTheoretical, chen2024mrpp, saparov2023cotYes, prabhakar2024cotYes, prystawski2023cotYes}, establishing it as a robust strategy for performance improvement.

However, despite their widespread success, CoT prompting faces notable limitations \citep{sprague2024cotNo}. 
Some studies argue that LLMs do not truly learn the underlying algorithms for solving problems \citep{stechly2024thoughtlessness} and often fail to faithfully explain the reasoning behind their answers \citep{turpin2023cotFaith}. 
By requiring LLMs to generate additional content, such as reasoning steps, CoT prompting may alter the internal states of the LLM \citep{feng2023cotTheoretical}. 
This raises an intriguing question: could this state alteration impact tasks that rely on these internal states, such as hallucination detection?

\subsection{Hallucination Detection}

Hallucinations in LLMs are commonly attributed to a discrepancy between the LLM's internal knowledge and real-world knowledge \citep{huang2024haluSurvey, zhang2024snowball}.
Some studies further suggest that hallucinations generated in earlier stages of text generation may propagate, leading to subsequent hallucinations \citep{zhang2024snowball, stechly2024thoughtlessness}. 
This observation aligns with our hypothesis that CoT prompting could influence hallucination detection by altering the LLM's internal states.

Our findings focus on a different aspect: 
the impact of CoT prompting on hallucination detection. 
Specifically, even in cases where hallucinations do not result from the propagation of earlier errors, hallucination detection methods may exhibit different performance under CoT prompting compared to standard prompting strategies.
This indicates that the additional reasoning steps introduced by CoT prompting may modify the internal states of model in ways that undermine the effectiveness of existing detection methods.

Methods for identifying and mitigating hallucinations can generally be classified based on whether they rely on external data.
Among methods that do not rely on external data, three main subcategories are identified: \textbf{consistency-based}, \textbf{internal-state-based}, and \textbf{self-evaluation-based} hallucination detection.

Internal-state-based methods utilize the LLM's internal representations or uncertainty during text generation to distinguish between hallucinated and non-hallucinated content. 
For instance, LLM-check \citep{sriramanan2024llmcheck} assesses variations in token probabilities during text generation.
Other approaches, such as DoLa \citep{chuang2024dola} and In-context Sharpness \citep{chen2024sharpness}, analyze features extracted from hidden states across different layers or time steps during decoding. 
These insights are then applied to detect and mitigate hallucinations.

Self-evaluation-based methods are founded on the premise that LLMs are capable of verbally evaluating the content they generate \citep{yao2023tot, wang2023selfconsist, yao2024got, kumar2024verbConfidence, diao2024active}.
These methods employ prompts to extract the LLM's self-evaluated confidence in its outputs.
By treating the LLM's verbalized certainty as an indicator of correctness, these methods identify hallucinations using confidence scores.

Consistency-based methods, such as SelfCheckGPT, operate under the assumption that if an LLM has accurate knowledge and confidence in its answers, it will generate semantically consistent outputs across multiple sampling attempts \citep{wang2023selfconsist, manakul2023selfcheck, ran2025divide}.
This approach is particularly relevant in factual QA tasks, where semantic consistency among sampled answers serves as an indicator of factuality. 
Inconsistent outputs, on the other hand, suggest the presence of hallucinations.

These three approaches are not mutually exclusive and can be combined. 
For example, INSIDE \citep{chen2024inside} integrates internal-state-based and consistency-based methods, while SelfCheckGPT \citep{manakul2023selfcheck} incorporates both self-evaluation-based and consistency-based techniques.

In contrast, hallucination detection methods that depend on external data are further categorized into retrieval-based and model-training-based approaches.
Retrieval-based methods utilize external knowledge bases to validate generated outputs \citep{min2023factscore, sankararaman2024provenance, li2024factChecking}. 
However, these methods often face limitations, including reliance on the quality and scope of external data, as well as the complexity of integrating retrieval and comparison systems.
Model-training-based methods, such as SAPLMA \citep{azaria2023saplma} and Inference-Time Intervention (ITI) \citep{li2023iti}, necessitate additional labeled data to train auxiliary models for hallucination detection.
These approaches, while effective, are resource-intensive and often lack generalization ability. 
For instance, ITI not only identifies hallucinations but also employs trained probes to address them. 
Nevertheless, the high resource demands and restricted plug-and-play functionality render these methods less practical for direct application to existing LLMs.

In light of these challenges, our study concentrates on hallucination detection methods that do not rely on external data, with a particular emphasis on consistency-based, internal-state-based, and self-evaluation-based approaches.
These methods provide enhanced flexibility and can be seamlessly integrated into existing LLM architectures without incurring additional resource overhead.

\section{Descriptions of CoT prompting}
\label{apd:prompts}

\textbf{CoT} encourages LLMs to think through problems step by step, fostering logical consistency and improving performance in complex tasks.
In our experiments, we use the following adapted template: \texttt{Think about it step by step.}

\textbf{LtM} guides the LLM to decompose complex questions into simpler subproblems, solve them individually, and recombine the answers to address the original question. 
The template used in our study is: \texttt{What subproblems must be solved before answering the inquiry?}

\textbf{MRPP} aims to minimize cognitive load by ensuring that each reasoning step is as simple and error-free as possible. 
The adapted template we employ is: \texttt{You need to perform multi-step reasoning, with each step carrying out as many basic operations as possible.}

Given these challenges, our study focuses on hallucination detection methods that do not require external data, particularly consistency-based, internal-state-based, and self-evaluation-based approaches.
These methods offer greater flexibility and can be readily applied to existing LLM architectures without additional resource overhead.

\section{Descriptions of Hallucination Detection Methods}
\label{apd:halu-detc}

\textbf{Perplexity} serves as a metric for assessing the predictive accuracy of an LLM on a given sequence.
It is defined as: 
\begin{equation}
PPL(x) = 2^{-\frac{1}{n} \sum_{i=1}^{n} \log P(x_i | x_{<i})}
\end{equation}
where $ x = \{x_1, x_2, \dots, x_n\} $ denotes the token sequence, and $ P(x_i | x_{<i}) $ indicates the conditional probability of token $ x_i $ given the preceding tokens. 

In a well-trained language model, sequences that conform to its learned language patterns are assigned higher probabilities, thereby yielding lower perplexity. 
By contrast, hallucinated content frequently diverges from the statistical patterns present in the training data, resulting in lower token probabilities and higher perplexity.
This renders perplexity a straightforward yet effective indicator for hallucination detection.

\textbf{Sharpness} \citep{chen2024sharpness} represents an internal-state-based method designed to assess the concentration of a LLM's hidden state during next-token prediction.
The method introduces a metric termed \textit{contextual entropy}, which measures the dispersion of the LLM's internal activations across input context tokens.
For a predicted token $v_p$, the contextual entropy is computed as:
\begin{equation}
\begin{split}
 E(v_p, v_{1:t}) & = -\sum_{i=1}^{t} \tilde{P}(activation=i \mid v_p, v_{1:t}) \\
& \times \log \tilde{P}(activation=i \mid v_p, v_{1:t}).
\end{split}
\end{equation}
Here, $\tilde{P}(\text{activation}=i \mid v_p, v_{1:t})$ denotes the normalized activation probability of token $v_p$ relative to each context token $v_i$. 
Lower entropy values signify more concentrated activations, implying sharper internal representations and greater confidence in token predictions.

In its original formulation, contextual entropy serves to adjust the original token probability distribution via a weighting coefficient.
This adjustment enables the method to regulate the influence of entropy on token probabilities, thereby improving its capacity to detect factually incorrect predictions.
By incorporating this entropy-based adjustment, the method successfully identifies discrepancies between high-confidence predictions and factually grounded content, rendering it well-suited for hallucination detection.

\textbf{Verbalized Certainty} \citep{kumar2024verbConfidence} is a self-evaluation-based method that enables the LLM to explicitly express its confidence level in the accuracy of its outputs. This is achieved by generating responses to a new query.
The method employs prompts to elicit the LLM's self-evaluation of response confidence, utilizing techniques such as \textit{Confidence Querying Prompts}, \textit{Simulation of a Third-Person Perspective}, and \textit{Likert Scale Utilization}.

\textbf{INSIDE} \citep{chen2024inside} represents a hybrid method that integrates internal-state-based and consistency-based approaches.
It assesses self-consistency across multiple responses by calculating the \textbf{EigenScore}, which measures semantic diversity in the embedding space. 
For $ K $ generated sequences, the covariance matrix of their sentence embeddings is computed as: 
\begin{equation}
\bm{\Sigma} = \mathbf{Z}^\top \cdot \mathbf{J}_d \cdot \mathbf{Z} 
\end{equation}
where $ \bm{\Sigma} \in \mathbb{R}^{K \times K} $ is the covariance matrix, $ \mathbf{Z} = [\bm{z}_1, \bm{z}_2, \dots, \bm{z}_K] \in \mathbb{R}^{d \times K} $ represents the embedding matrix, and $ \mathbf{J}_d = \bm{I}_d - \frac{1}{d} \mathbf{1}_d \mathbf{1}_d^\top $ is the centering matrix.  
The EigenScore is defined as the log-determinant of the regularized covariance matrix: 
\begin{equation}
E(\mathcal{Y}|\bm{x}, \bm{\theta}) = \frac{1}{K} \log \det(\bm{\Sigma} + \alpha \cdot \mathbf{I}_K)
\end{equation}
where $ \alpha \cdot \mathbf{I}_K $ ensures numerical stability, and the eigenvalues $ \lambda_i $  are utilized to compute: 
\begin{equation}
E(\mathcal{Y}|\bm{x}, \bm{\theta}) = \frac{1}{K} \sum_{i=1}^K \log(\lambda_i).
\end{equation}
This metric effectively captures the self-consistency of generated responses and proves particularly adept at identifying hallucinations by detecting inconsistencies across multiple samples.

\textbf{SelfCheckGPT} \citep{manakul2023selfcheck} utilizes consistency-based methods, including Natural Language Inference (NLI) and self-evaluation-based techniques, to evaluate hallucinations. 
In its NLI-based form, SelfCheckGPT compares LLM-generated responses with stochastically generated alternatives. It then categorizes the relationships between the hypothesis (response) and premise (context) into three categories: \textit{entailment}, \textit{neutral}, or \textit{contradiction}.
The contradiction score serves as the SelfCheckGPT score.
Alternatively, in its self-evaluation-based form, SelfCheckGPT directly queries the LLM to verify whether specific sentences are supported by the generated context through carefully designed prompts.

\textbf{LLM-Check} \citep{sriramanan2024llmcheck} represents an internal-state-based method that assesses hallucinations by analyzing the covariance structure of token embeddings within a single response.
Similar to INSIDE's EigenScore, LLM-Check computes the covariance matrix of hidden representations: 
\begin{equation}
\bm{\Sigma}_2 = \mathbf{H}^\top \mathbf{H}
\end{equation}
and defines the hallucination score as the mean log-determinant of $ \bm{\Sigma}_2 $: 
\begin{equation}
\text{Score} = \frac{2}{m} \sum_{i=1}^m \log \sigma_i
\end{equation}
where $ \sigma_i $ are the singular values of $ \mathbf{H} $.

\section{Details on the Evaluation Framework}
\label{apd:framework}
The evaluation framework aims to systematically examine the relationship between hallucination scores and correctness under both CoT prompting and non-CoT prompting conditions.
The framework includes the following components:

\textbf{Distribution Consistency Test.} We use statistical methods, such as the Kolmogorov-Smirnov test, to compare the distribution of hallucination scores with and without CoT prompting.
This analysis determines whether the introduction of CoT prompting significantly alters the distribution of hallucination scores. 
It also provides an intuitive measure of how CoT prompting influences hallucination detection.

\textbf{Classification Performance.} We treat correctness as the ground truth (binary: correct or incorrect) and hallucination scores as predictions. 
We then use AUROC to quantify the performance of hallucination detection methods before and after applying CoT prompting.
This metric quantifies the degree to which CoT prompting influences the effectiveness of hallucination detection methods.

\textbf{Monotonic Relationship.} Using Spearman's rank correlation coefficient, we evaluate the monotonic relationship between hallucination scores and correctness values. 
Here, correctness is treated as a continuous variable rather than a binary label. 
This metric shows whether hallucination detection scores are consistent with correctness under CoT prompting.

\textbf{Alignment.} We use Expected Calibration Error (ECE) to assess how well hallucination scores align with correctness.
By treating correctness as the ground truth and hallucination scores as predictions, this metric evaluates whether hallucination detection methods effectively reflect the degree of correctness. 
This approach goes beyond simply distinguishing between correct and incorrect responses.
This is particularly important as some answers may not be strictly correct or incorrect, and well-calibrated hallucination scores should reflect the nuances of correctness.

This evaluation framework collectively enables a comprehensive analysis of whether CoT prompting confuses hallucination detection methods. 
By observing performance degradation, changes in distribution, and alignment issues, our methodology provides a systematic approach to understanding and quantifying the impact of reasoning steps on hallucination detection.

\section{Experiments Implementation Details}
\label{apd:details}

\subsection{Pilot Experiment}

The normalization process for token probabilities in candidate options can be mathematically described as follows:
\begin{equation}
{p}_{\text{o}} = \frac{{p}_{\text{t}}[{t}_{\text{o}}]}{\sum_{i=1}^{N} {p}_{\text{t}}[{t}_{i}]}, 
\hat{y} = \arg\max_{i} {p}_{\text{o}}[i],
\end{equation}
where ${t}_{\text{o}}$ and ${p}_{\text{o}}$ denote the tokens and normalized probabilities of each candidate option, respectively. 
$N$ is the total number of candidate options, and $\hat{y}$ is the LLM's predicted option, determined by selecting the one with the highest probability. 
The predicted answer $\hat{y}$ is then compared to the ground-truth answer to determine whether the LLM's response is correct. 
A correct prediction implies no hallucination, while an incorrect one corresponds to a potential hallucination.

For experiments involving CoT prompting, we first prompt the LLM to generate reasoning steps and then append a final query about the candidate options to determine the final answer.
For the baseline (direct answer) setup, we directly prompt the LLM for the final answer without generating reasoning steps. 
We generate reasoning steps using the \texttt{vLLM}\footnote{\url{https://docs.vllm.ai/en/v0.6.1.post1/}} with \texttt{temperature} = 0.5 and \texttt{max\_tokens} = 512.

Once the complete prompt is constructed, we feed it into the LLM to compute token-level probabilities for the candidate options.
These probabilities are recorded for subsequent calculations of accuracy, entropy, and AUROC.

By systematically comparing LLMs' performance with and without CoT prompting, we aim to evaluate their ability to improve confidence and accuracy, as well as their impact on hallucination detection.

\subsection{Main Experiment}

We conduct all experiments using four open-source instruction-tuned LLMs: Llama-3.1-8B-Instruct\footnote{\url{https://huggingface.co/meta-llama/Llama-3.1-8B-Instruct}}: A lightweight yet capable language model fine-tuned for instruction-following tasks, balancing efficiency and performance in resource-constrained environments \citep{dubey2024llama}.  Mistral-7B-v0.3-Instruct\footnote{\url{https://huggingface.co/mistralai/Mistral-7B-Instruct-v0.3}}: A compact model optimized for competitive reasoning and low-latency inference, trained via advanced instruction tuning \citep{jiang2023mistral}.  Llama-3.1-70B-Instruct\footnote{\url{https://huggingface.co/meta-llama/Llama-3.1-70B-Instruct}}: A 70-billion-parameter Transformer-based model designed for high-precision instruction alignment, excelling in complex reasoning and human-preference tasks across diverse domains \citep{dubey2024llama}.  DeepSeek-R1-Distill-Llama-8B\footnote{\url{https://huggingface.co/deepseek-ai/DeepSeek-R1-Distill-Llama-8B}}: An 8-billion-parameter distilled variant leveraging knowledge transfer from the 671B DeepSeek-R1 model, specialized in step-by-step mathematical reasoning and efficient deployment on consumer-grade hardware \citep{guo2025deepseek}.  

Unless otherwise specified, all generative processes are conducted using the \texttt{vLLM} inference framework with a temperature setting of $ T = 0.5 $.
The system prompt used for response generation was: \texttt{You are a helpful assistant.} For the reasoning-oriented model, we first force the output of \texttt{<think>\textbackslash n\textbackslash n</think>} and then let the model continue generating, thereby implementing the ``without CoT'' setting.

When generating reasoning content, we apply the corresponding prompts discussed in Section~\ref{subsec:prompts}.
For consistency-based hallucination detection methods such as INSIDE and SelfCheckGPT, which involve stochastic sampling, we adopt the generation parameters reported in their respective papers.
For INSIDE, we set \texttt{n=15, temperature=0.5, top\_p=0.99,} and \texttt{top\_k=5}. For SelfCheckGPT, the parameters are \texttt{n=20} and \texttt{temperature=1}.

For internal-state-based hallucination detection methods, including INSIDE, Sharpness, and LLM-Check, we determine layer-specific settings based on the best-performing configurations reported in the literature.
Specifically, for INSIDE, we use layer 17; for Sharpness, layer 26; and for LLM-Check, we use layer 15 for the Hidden Score and layer 23 for the Attention Score.
For Sharpness, which involves adjusting the entropy's impact on the original token distribution using a parameter $\alpha$, we set $\alpha=1.0$ according to the settings in its original paper.

TruthfulQA offers a training set with 6.9k examples, which we use to fine-tune the \texttt{GPT-3-6.7B} model for evaluation purposes.
Since OpenAI's API no longer provides access to this model, we instead fine-tune the \texttt{Llama-3.1-8B}\footnote{\url{https://huggingface.co/meta-llama/Llama-3.1-8B}} model using LoRA \citep{hu2022lora} to compute the Truthfulness and Informativeness scores.
We conduct fine-tuning on the \texttt{LLaMA-Factory}\footnote{\url{https://github.com/hiyouga/LLaMA-Factory}} platform with the following hyperparameters: \texttt{learning rate = 5.0e-05, batch size = 32, epochs = 5, lora\_rank = 8}, and \texttt{lora\_alpha = 16}.

The evaluation metrics are defined in Section \ref{subsec:eval}. 
For Expected Calibration Error (ECE), we apply a normalization process to ensure predictions fall within the $[0, 1]$ range. 
Specifically, we scale the data within the $3\%-97\%$ range to $[0, 1]$, while outliers beyond this range are clipped to $0$ or $1$.
This normalization avoids distortions caused by extreme values. 
Verbalized Certainty is excluded from this analysis because its hallucination detection scores are confined to five distinct levels.

\section{Results}
\label{apd:results}

\subsection{MIND Evaluation Details}

To explore whether our observations extend to model-training-based detection paradigms, we conducted additional experiments using the MIND framework \citep{su2024mind}.

MIND is an unsupervised, internal-state-based hallucination detection framework. Its key advantages include:

\begin{itemize}
    \item Leveraging pseudo-labeled data automatically constructed from Wikipedia for training, thus avoiding manual annotation.
    \item Training lightweight classifiers on the LLM's hidden states for real-time detection.
\end{itemize}
However, it also shares limitations with other model-training-based approaches:
\begin{itemize}
    \item It requires training an auxiliary detector tailored to each specific target LLM.
    \item Its performance is dependent on the design and quality of the pseudo-labeled data.
\end{itemize}
This need for model-specific training is why we primarily focused on non-training-based detection methods in our main experiments.

\textbf{Experimental Setup.} Due to the tight coupling between MIND's released code and its original LLM architecture, our evaluation was limited to the \textbf{LLaMA-2-7b-hf} model. We evaluated its performance on the \textbf{HaluEval} dataset under different prompting scenarios: standard prompting (\textit{base}), and the three CoT prompting methods studied in this work.

\textbf{Results.} The results, presented in Table~\ref{tab:mind-results}, show that MIND's detection performance also degrades when the LLM responds after CoT prompting. This degradation is consistent across both \textit{Textual} and \textit{Semantic} hallucination annotations and across both AUROC and Correlation metrics, reinforcing the broad and challenging nature of the phenomenon we identify.

\begin{table}[h]
\centering

\caption{Performance of the MIND detector on HaluEval under different prompting scenarios.}
\label{tab:mind-results}
\begin{tabular}{l|cc|cc}
\toprule
\multirow{2}{*}{} & \multicolumn{2}{c|}{Textual} & \multicolumn{2}{c}{Semantic} \\
 & AUROC & Corr & AUROC & Corr \\ 
 \midrule
base & 0.6765 & 0.4733 & 0.5904 & 0.2941 \\
CoT & 0.6521 & 0.4261 & 0.5572 & 0.2307 \\
MRPP & 0.6513 & 0.4304 & 0.5608 & 0.2381 \\
LtM & 0.6379 & 0.3983 & 0.5573 & 0.2246 \\ 
\bottomrule
\end{tabular}
\end{table}

\subsection{Detail of Results}

\begin{figure*}[]
  \centering
  \includegraphics[width=0.9\textwidth]{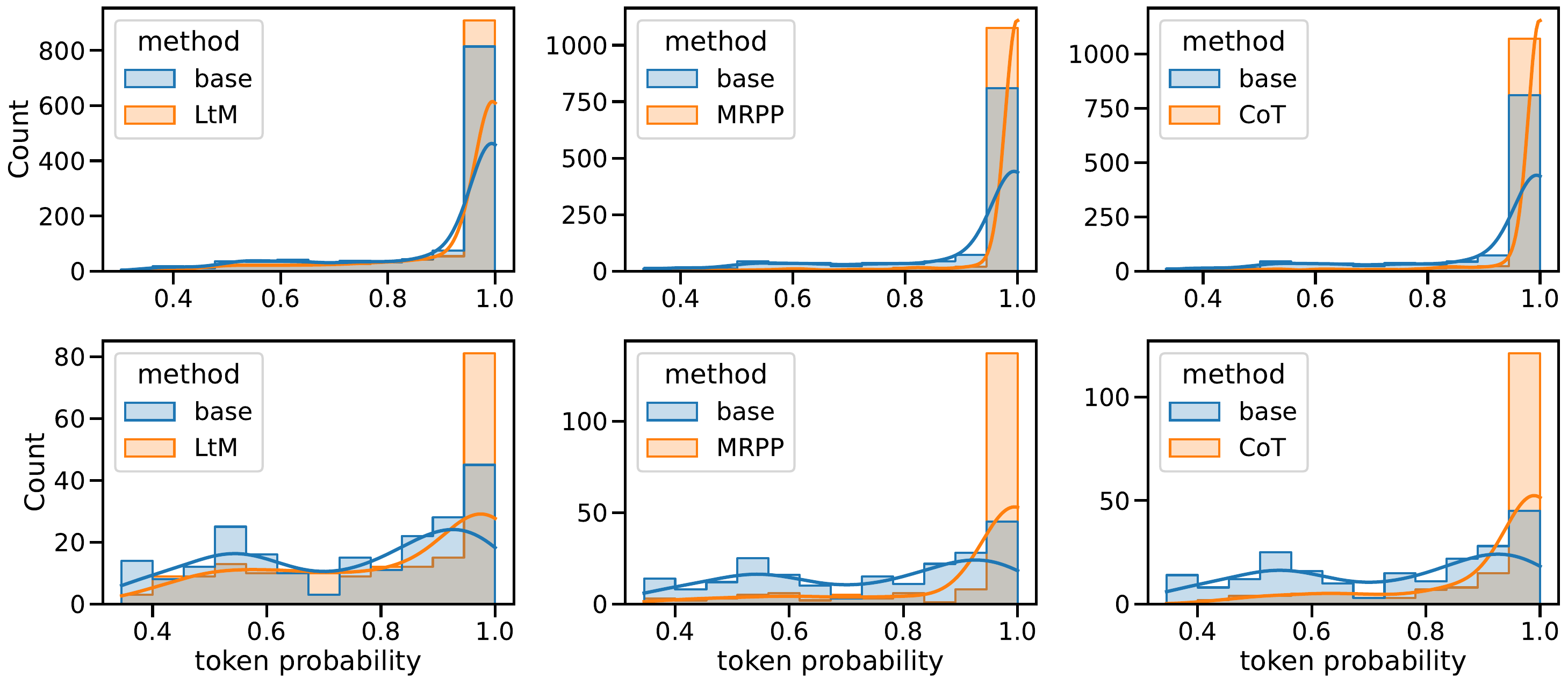}
  \caption{Comparison of token probability distributions for \texttt{Llama-3.1-8B-Instruct} on ARC-Challenge. The upper row shows all samples; the lower row shows incorrect predictions after CoT prompting. Kolmogorov-Smirnov test confirms significant differences ($p < 0.01$) between conditions with and without CoT prompting. }
  \label{fig:token-level-llama}
\end{figure*}

\input{tables/tqa-acc}
\input{tables/auroc-llama}

\input{tables/auroc-mistral}
\input{tables/auroc-r1}
\input{tables/auroc-70b}
\input{tables/auroc-cnn}

\input{tables/corr-llama}
\input{tables/corr-mistral}
\input{tables/corr-r1}
\input{tables/corr-70b}
\input{tables/corr-cnn}
\input{tables/ece-llama}
\input{tables/ece-mistral}
\input{tables/ece-r1}
\input{tables/ece-70b}
\input{tables/ece-cnn}
\input{tables/tqa-roc-full}
\input{tables/qwen-judge}

%% file: tables/tqa-acc.tex
% Table generated by Excel2LaTeX from sheet 'truthfulqa'
\begin{table}[htbp]
  \centering
  \caption{Results of different LLMs and CoT prompting methods on the TruthfulQA dataset (values ×100). All metrics are positively oriented. Improvements after CoT prompting are bolded. Notably, the Reasoning model shows decreased Truthfulness but increased Informativeness after reasoning.}
\begin{tabular}{lrr}
    \toprule
      & \multicolumn{1}{l}{Truthfulness} & \multicolumn{1}{l}{Informativeness} \\
    \midrule
    \multicolumn{3}{c}{Llama-3.1-8B-Instruct} \\
    base & 65.73  & 94.37  \\
    CoT & \textbf{65.85} & \textbf{97.18} \\
    LtM & \textbf{74.17} & \textbf{97.55} \\
    MRPP & 52.63  & \textbf{95.23} \\
    \midrule
    \multicolumn{3}{c}{Mistral-7B-Instruct-v0.3} \\
    base & 85.68  & 93.64  \\
    CoT & \textbf{87.88} & \textbf{97.06} \\
    LtM & 80.29  & \textbf{95.72} \\
    MRPP & 83.23  & \textbf{96.21} \\
    \midrule
    \multicolumn{3}{c}{DeepSeek-R1-Distill-Llama-8B} \\
    base & 56.18  & 94.86  \\
    CoT & 21.18  & \textbf{97.92} \\
    LtM & 25.83  & \textbf{96.45} \\
    MRPP & 22.77  & \textbf{96.82} \\
    \bottomrule
    \end{tabular}%
  \label{tab:tqa-acc}%
\end{table}%

%% file: tables/auroc-llama.tex
% Table generated by Excel2LaTeX from sheet 'roc'
\begin{table*}[htb]
  \centering
  \small
  \caption{AUROC results of \texttt{Llama-3.1-8B-Instruct} across different datasets. Bolded values indicate cases with reduced performance.}
    \begin{tabular}{cccccccccc}
    \toprule
      &   & PPL & Sharpness & \makecell{Eigen\\Score} & \makecell{SelfCk\\-Prompt} & \makecell{SelfCk\\-NLI} & \makecell{Verbalized\\Certainty} & \makecell{Hidden\\Score} & \makecell{Attn\\Score} \\
      \midrule
    \multicolumn{1}{c}{\multirow{4}[1]{*}{\makecell{Textual \\ HaluEval}}} & base & 65.77  & 65.67  & 60.75  & 69.58  & 62.34  & 75.37  & 53.07  & 58.65  \\
      & CoT & 67.89  & 67.42  & 64.33  & \textbf{61.93} & 69.14  & \textbf{62.47} & 54.51  & \textbf{53.79} \\
      & LtM & 67.77  & 67.59  & 63.04  & \textbf{62.75} & 67.79  & \textbf{62.91} & \textbf{51.62} & \textbf{56.10} \\
      & MRPP & 67.48  & 67.22  & 66.73  & \textbf{62.82} & 67.75  & \textbf{61.54} & 57.34  & \textbf{56.46} \\
    \midrule
    \multicolumn{1}{c}{\multirow{4}[2]{*}{\makecell{Semantic \\ HaluEval}}} & base & 59.76  & 59.62  & 62.44  & 58.34  & 55.68  & 65.08  & 62.86  & 57.01  \\
      & CoT & 64.12  & 64.95  & \textbf{61.33} & \textbf{53.78} & \textbf{50.81} & 69.59  & \textbf{61.54} & \textbf{54.14} \\
      & LtM & 61.14  & 61.26  & \textbf{61.92} & \textbf{54.29} & 58.92  & \textbf{59.71} & \textbf{56.84} & 57.97  \\
\bottomrule
    \end{tabular}%
  \label{tab:auroc-llama}%
\end{table*}

%% file: tables/auroc-mistral.tex
\begin{table*}[htb]
  \centering
  \small
  \caption{AUROC results of \texttt{Mistral-7B-Instruct-v0.3} across different datasets. The bolded values indicate cases with reduced performance.}
   \begin{tabular}{clcccccccc}
    \toprule
      &   & PPL & Sharpness & \makecell{Eigen\\Score} & \makecell{SelfCk\\-Prompt} & \makecell{SelfCk\\-NLI} & \makecell{Verbalized\\Certainty} & \makecell{Hidden\\Score} & \makecell{Attn\\Score} \\
    \midrule
    \multicolumn{1}{c}{\multirow{4}[2]{*}{\makecell{Textual \\ HaluEval}}} & base & 65.90  & 65.89  & 63.45  & 76.71  & 67.24  & 57.13  & 56.49  & 58.96  \\
      & CoT & \textbf{64.96} & \textbf{64.96} & 64.24  & \textbf{70.47} & \textbf{63.12} & 59.08  & \textbf{55.90} & 60.55  \\
      & LtM & \textbf{63.94} & \textbf{63.94} & 64.69  & \textbf{68.64} & \textbf{63.24} & 61.30  & \textbf{56.27} & 59.73  \\
      & MRPP & \textbf{64.18} & \textbf{64.18} & 64.96  & \textbf{69.11} & \textbf{59.26} & 62.46  & 59.47  & 61.39  \\
    \midrule
    \multicolumn{1}{c}{\multirow{4}[2]{*}{\makecell{Semantic \\ HaluEval}}} & base & 63.41  & 63.41  & 63.49  & 59.39  & 58.04  & 63.39  & 57.88  & 60.93  \\
      & CoT & 75.51  & 75.56  & \textbf{63.05} & \textbf{52.94} & \textbf{51.29} & 75.94  & \textbf{50.65} & 65.49  \\
      & LtM & 72.70  & 72.60  & 79.83  & \textbf{58.57} & \textbf{56.60} & 80.34  & \textbf{52.94} & 63.20  \\
\bottomrule
   \end{tabular}%
  \label{tab:auroc-mistral}%
 \end{table*}%

%% file: tables/auroc-r1.tex
\begin{table*}[htbp]
  \centering
  \small
  \caption{AUROC results of \texttt{DeepSeek-R1-Distill-Llama-8B} across different datasets. The bolded values indicate cases with reduced performance.}
    \begin{tabular}{cccccccccc}
    \toprule
      &   & \multicolumn{1}{c}{PPL} & \multicolumn{1}{c}{Sharpness} & \makecell{Eigen\\Score} & \makecell{SelfCk\\-Prompt} & \makecell{SelfCk\\-NLI} & \makecell{Hidden\\Score} & \makecell{Verbalized\\Certainty} & \makecell{Attn\\Score} \\
    \midrule
    \multicolumn{1}{c}{\multirow{4}[1]{*}{\makecell{Textual \\ HaluEval}}} & base & 0.5287  & 0.5281  & 0.6097  & 0.6464  & 0.6543  & 0.6094  & 0.6094  & 0.6346  \\
      & CoT & \textbf{0.5142} & \textbf{0.5161} & \textbf{0.5192} & \textbf{0.6450} & \textbf{0.5932} & \textbf{0.5492} & \textbf{0.5492} & \textbf{0.6156} \\
      & LtM & \textbf{0.5063} & \textbf{0.5089} & \textbf{0.5202} & \textbf{0.6276} & \textbf{0.5895} & \textbf{0.5117} & \textbf{0.5117} & 0.6502  \\
      & MRPP & \textbf{0.5135} & \textbf{0.5129} & \textbf{0.5274} & \textbf{0.6313} & \textbf{0.5773} & \textbf{0.5338} & \textbf{0.5338} & \textbf{0.6258} \\
    \midrule
    \multicolumn{1}{c}{\multirow{4}[1]{*}{\makecell{Semantic \\ HaluEval}}} & base & 0.6648  & 0.6643  & 0.6543  & 0.5033  & 0.5207  & 0.5382  & 0.5382  & 0.7279  \\
      & CoT & \textbf{0.6546} & \textbf{0.6541} & \textbf{0.5513} & \textbf{0.5003} & 0.5493  & \textbf{0.5061} & \textbf{0.5061} & \textbf{0.5058} \\
      & LtM & \textbf{0.6571} & \textbf{0.6563} & \textbf{0.5830} & 0.5491  & 0.5434  & \textbf{0.5019} & \textbf{0.5019} & \textbf{0.5136} \\
\bottomrule
    \end{tabular}%
  \label{tab:auroc-r1}%
\end{table*}%

%% file: tables/auroc-70b.tex
% Table generated by Excel2LaTeX from sheet 'roc'
\begin{table*}[htbp]
  \centering
  \small
  \caption{AUROC results of \texttt{Llama-3.1-70B-Instruct} across different datasets. Bolded values indicate cases with reduced performance.}
    \begin{tabular}{ccrrrrrrrr}
    \toprule
      &   & \multicolumn{1}{c}{PPL} & \multicolumn{1}{c}{Sharpness} & \makecell{Eigen\\Score} & \makecell{SelfCk\\-Prompt} & \makecell{SelfCk\\-NLI} & \makecell{Verbalized\\Certainty} & \makecell{Hidden\\Score} & \makecell{Attn\\Score} \\
    \midrule
    \multicolumn{1}{c}{\multirow{4}[2]{*}{\makecell{Textual \\ HaluEval}}} & base & 0.7561  & 0.7561  & 0.7311  & 0.6877  & 0.6923  & 0.6182  & 0.7303  & 0.5721  \\
      & CoT & \textbf{0.6817} & \textbf{0.6802} & 0.7694  & 0.7014  & \textbf{0.6018} & \textbf{0.5503} & \textbf{0.6390} & 0.7005  \\
      & LtM & \textbf{0.7223} & \textbf{0.7226} & 0.7423  & \textbf{0.6503} & \textbf{0.6232} & \textbf{0.5542} & \textbf{0.6725} & 0.6847  \\
      & MRPP & \textbf{0.7491} & \textbf{0.7493} & 0.7762  & 0.6877  & \textbf{0.6250} & \textbf{0.5527} & 0.7451  & 0.6724  \\
    \midrule
    \multicolumn{1}{c}{\multirow{4}[2]{*}{\makecell{Semantic \\ HaluEval}}} & base & 0.7241  & 0.7234  & 0.6998  & 0.5782  & 0.5997  & 0.5938  & 0.6019  & 0.5165  \\
      & CoT & \textbf{0.6870} & \textbf{0.6870} & \textbf{0.6527} & 0.6226  & \textbf{0.5514} & \textbf{0.5275} & 0.6782  & 0.5889  \\
      & LtM & \textbf{0.7130} & \textbf{0.7128} & 0.7123  & 0.5816  & \textbf{0.5465} & \textbf{0.5664} & 0.6509  & 0.5980  \\
\bottomrule
    \end{tabular}%
  \label{tab:auroc-70b}%
\end{table*}%

%% file: tables/auroc-cnn.tex
\begin{table*}[htbp]
  \centering
  \small
  \caption{AUROC results of CNN/Daily Mail dataset across different LLMs. Bolded values indicate cases with reduced performance.}
    \begin{tabular}{cccccccccc}
    \toprule
      &   & \multicolumn{1}{c}{PPL} & \multicolumn{1}{c}{Sharpness} & \makecell{Eigen\\Score} & \makecell{SelfCk\\-Prompt} & \makecell{SelfCk\\-NLI} & \makecell{Hidden\\Score} & \makecell{Verbalized\\Certainty} & \makecell{Attn\\Score} \\
    \midrule
    \multicolumn{10}{c}{Llama-3.1-8B-Instruct} \\
    \multicolumn{1}{c}{\multirow{4}[1]{*}{Textual}} & base & 0.5891  & 0.5861  & 0.6295  & 0.5351  & 0.5877  & 0.5139  & 0.5398  & 0.6332  \\
      & CoT & \textbf{0.5789} & \textbf{0.5791} & \textbf{0.6059} & \textbf{0.5001} & \textbf{0.5701} & 0.5386  & \textbf{0.5310} & \textbf{0.6049} \\
      & LtM & 0.6267  & 0.6240  & 0.6444  & \textbf{0.5188} & \textbf{0.5376} & 0.5425  & \textbf{0.5354} & \textbf{0.6138} \\
      & MRPP & 0.6049  & 0.5982  & 0.6228  & \textbf{0.5033} & \textbf{0.5517} & 0.5335  & 0.5409  & \textbf{0.6040} \\
    \midrule
    \multicolumn{1}{c}{\multirow{4}[2]{*}{Semantic}} & base & 0.5681  & 0.5704  & 0.6395  & 0.5260  & 0.5610  & 0.5199  & 0.5287  & 0.6695  \\
      & CoT & \textbf{0.5531} & \textbf{0.5510} & \textbf{0.5957} & 0.5278  & \textbf{0.5258} & \textbf{0.5144} & 0.5320  & \textbf{0.6625} \\
\bottomrule
    \end{tabular}%
  \label{tab:auroc-cnn}%
\end{table*}%

%% file: tables/corr-llama.tex
\begin{table*}[htb]
  \centering
  \small
  \caption{Spearman correlation results of \texttt{Llama-3.1-8B-Instruct} across different datasets. The bolded values indicate cases with reduced correlation.}
    \begin{tabular}{clcccccccc}
    \toprule
      &   & PPL & Sharpness & \makecell{Eigen\\Score} & \makecell{SelfCk\\-Prompt} & \makecell{SelfCk\\-NLI} & \makecell{Verbalized\\Certainty} & \makecell{Hidden\\Score} & \makecell{Attn\\Score} \\
    \midrule
    \multirow{4}[2]{*}{Textual} & base & -25.10  & -24.94  & -17.11  & 31.15  & -19.64  & 46.46  & 4.89  & 13.77  \\
      & CoT & -29.85  & -29.07  & -23.91  & \textbf{19.91} & -31.93  & \textbf{25.68} & 7.52  & \textbf{6.32} \\
      & LtM & -29.72  & -29.42  & -21.81  & \textbf{21.33} & -29.75  & \textbf{26.53} & \textbf{2.71} & \textbf{10.20} \\
      & MRPP & -29.07  & -28.64  & -27.82  & \textbf{21.32} & -29.53  & \textbf{24.15} & 12.21  & \textbf{10.74} \\
    \midrule
    \multirow{4}[2]{*}{Semantic} & base & -19.59  & -19.27  & -28.14  & 14.93  & -13.04  & 30.52  & 28.48  & 18.60  \\
      & CoT & -41.67  & -41.99  & -29.48  & \textbf{7.53} & \textbf{-10.86} & \textbf{29.39} & \textbf{27.54} & \textbf{6.48} \\
      & LtM & -21.30  & -21.53  & \textbf{-23.95} & \textbf{8.13} & -17.43  & \textbf{21.99} & \textbf{13.93} & \textbf{15.66} \\
\bottomrule
    \end{tabular}%
  \label{tab:corr-llama}%
\end{table*}%

%% file: tables/corr-mistral.tex
% Table generated by Excel2LaTeX from sheet 'corr'
\begin{table*}[htb]
  \centering
  \small
  \caption{Spearman correlation results of \texttt{Mistral-7B-Instruct-v0.3} across different datasets. The bolded values indicate cases with reduced correlation.}
    \begin{tabular}{clcccccccc}
    \toprule
      &   & PPL & Sharpness & \makecell{Eigen\\Score} & \makecell{SelfCk\\-Prompt} & \makecell{SelfCk\\-NLI} & \makecell{Verbalized\\Certainty} & \makecell{Hidden\\Score} & \makecell{Attn\\Score} \\
    \midrule
    \multicolumn{1}{c}{\multirow{4}[1]{*}{\makecell{Textual \\ HaluEval}}} & base & -27.09  & -27.08  & -22.92  & 45.51  & -29.37  & 16.40  & 11.06  & 15.26  \\
      & CoT & \textbf{-24.96} & \textbf{-24.97} & -23.76  & \textbf{34.16} & \textbf{-21.90} & 17.88  & \textbf{9.84} & 17.62  \\
      & LtM & \textbf{-23.03} & \textbf{-23.04} & -24.26  & \textbf{30.79} & \textbf{-21.87} & 19.41  & \textbf{10.35} & 16.08  \\
      & MRPP & \textbf{-23.08} & \textbf{-23.08} & -24.34  & \textbf{31.10} & \textbf{-15.07} & 21.79  & 15.42  & 18.54  \\
    \midrule
    \multicolumn{1}{c}{\multirow{4}[0]{*}{\makecell{Semantic \\ HaluEval}}} & base & -21.87  & -21.86  & -22.16  & 16.47  & -16.01  & 26.81  & 15.33  & 20.13  \\
      & CoT & -37.38  & -37.38  & -30.76  & 16.77  & \textbf{-9.00} & 38.16  & 16.26  & \textbf{1.65} \\
      & LtM & -55.59  & -55.59  & -53.57  & 34.42  & -24.33  & 45.13  & 28.04  & \textbf{18.37} \\
\bottomrule
    \end{tabular}%
  \label{tab:corr-mistral}%
\end{table*}

%% file: tables/corr-r1.tex
% Table generated by Excel2LaTeX from sheet 'corr'
\begin{table*}[htbp]
  \centering
  \small
  \caption{Spearman correlation results of \texttt{DeepSeek-R1-Distill-Llama-8B} across different datasets. The bolded values indicate cases with reduced correlation.}
    \begin{tabular}{ccrrrrrrr}
    \toprule
      &   & \multicolumn{1}{c}{PPL} & \multicolumn{1}{c}{Sharpness} & \makecell{Eigen\\Score} & \makecell{SelfCk\\-Prompt} & \makecell{SelfCk\\-NLI} & \makecell{Verbalized\\Certainty} & \makecell{Hidden\\Score} \\
    \midrule
    \multicolumn{1}{c}{\multirow{4}[2]{*}{\makecell{Textual \\ HaluEval}}} & base & -0.0378  & -0.0369  & -0.1444  & 0.1928  & -0.2031  & 0.1440  & 0.1772  \\
      & CoT & \textbf{-0.0224} & \textbf{-0.0253} & \textbf{-0.0302} & 0.2285  & \textbf{-0.1468} & \textbf{0.0775} & 0.1821  \\
      & LtM & \textbf{-0.0099} & \textbf{-0.0139} & \textbf{-0.0315} & 0.1994  & \textbf{-0.1398} & \textbf{0.0183} & 0.2347  \\
      & MRPP & \textbf{-0.0212} & \textbf{-0.0203} & \textbf{-0.0430} & 0.2062  & \textbf{-0.1213} & \textbf{0.0530} & 0.1975  \\
    \midrule
    \multicolumn{1}{c}{\multirow{4}[2]{*}{\makecell{Semantic \\ HaluEval}}} & base & -0.3112  & -0.3104  & -0.2987  & 0.0689  & -0.0066  & 0.0434  & 0.3913  \\
      & CoT & \textbf{-0.3107} & \textbf{-0.3040} & \textbf{-0.1378} & \textbf{0.0165} & -0.0120  & 0.0884  & \textbf{0.0135} \\
      & LtM & -0.3373  & -0.3267  & \textbf{-0.1399} & \textbf{0.0554} & -0.0557  & 0.0447  & \textbf{0.0510} \\
\bottomrule
    \end{tabular}%
  \label{tab:corr-r1}%
\end{table*}%

%% file: tables/corr-70b.tex
% Table generated by Excel2LaTeX from sheet 'corr'
\begin{table*}[htbp]
  \centering
  \small
  \caption{Spearman correlation results of \texttt{Llama-3.1-70B-Instruct} across different datasets. The bolded values indicate cases with reduced correlation.}
    \begin{tabular}{ccrrrrrrrr}
    \toprule
      &   & \multicolumn{1}{c}{PPL} & \multicolumn{1}{c}{Sharpness} & \makecell{Eigen\\Score} & \makecell{SelfCk\\-Prompt} & \makecell{SelfCk\\-NLI} & \makecell{Verbalized\\Certainty} & \makecell{Hidden\\Score} & \makecell{Attn\\Score} \\
    \midrule
    \multicolumn{1}{c}{\multirow{4}[2]{*}{\makecell{Textual \\ HaluEval}}} & base & -0.4222  & -0.4222  & -0.4078  & 0.3499  & -0.3632  & 0.2789  & 0.2739  & 0.0287  \\
      & CoT & \textbf{-0.3453} & \textbf{-0.3435} & -0.4282  & \textbf{0.3363} & \textbf{-0.1636} & \textbf{0.1409} & 0.2894  & \textbf{0.2698} \\
      & LtM & \textbf{-0.3669} & \textbf{-0.3673} & \textbf{-0.3753} & \textbf{0.2490} & \textbf{-0.1571} & \textbf{0.1589} & \textbf{0.2312} & \textbf{0.2532} \\
      & MRPP & \textbf{-0.3914} & \textbf{-0.3915} & \textbf{-0.4015} & \textbf{0.2812} & \textbf{-0.2027} & \textbf{0.1650} & \textbf{0.1693} & \textbf{0.2208} \\
    \midrule
    \multicolumn{1}{c}{\multirow{4}[2]{*}{\makecell{Semantic \\ HaluEval}}} & base & -0.4651  & -0.4638  & -0.4184  & 0.1525  & -0.2041  & 0.2288  & 0.1875  & 0.1358  \\
      & CoT & \textbf{-0.4201} & \textbf{-0.4201} & \textbf{-0.3420} & 0.2450  & \textbf{-0.1057} & \textbf{0.0682} & 0.3507  & 0.1896  \\
      & LtM & -0.4958  & -0.4955  & -0.4731  & 0.1732  & \textbf{-0.1472} & \textbf{0.1527} & 0.2608  & 0.2119  \\
\bottomrule
    \end{tabular}%
  \label{tab:corr-70b}%
\end{table*}%

%% file: tables/corr-cnn.tex
% Table generated by Excel2LaTeX from sheet 'corr'
\begin{table*}[htbp]
  \centering
  \small
  \caption{Spearman correlation results of CNN/Daily Mail dataset across different LLMs. The bolded values indicate cases with reduced correlation.}
    \begin{tabular}{ccrrrrrrr}
    \toprule
      &   & \multicolumn{1}{c}{PPL} & \multicolumn{1}{c}{Sharpness} & \makecell{Eigen\\Score} & \makecell{SelfCk\\-Prompt} & \makecell{SelfCk\\-NLI} & \makecell{Hidden\\Score} & \makecell{Attn\\Score} \\
    \midrule
    \multicolumn{9}{c}{Llama-3.1-8B-Instruct} \\
    \multicolumn{1}{c}{\multirow{4}[1]{*}{Textual}} & base & -0.1595  & -0.1591  & -0.2852  & -0.0788  & -0.1843  & -0.0274  & 0.2857  \\
      & CoT & -0.2116  & -0.2067  & \textbf{-0.2452} & \textbf{-0.0039} & \textbf{-0.1306} & -0.0777  & \textbf{0.2366} \\
      & LtM & -0.2725  & -0.2664  & -0.2993  & \textbf{-0.0438} & \textbf{-0.0853} & -0.0885  & \textbf{0.2423} \\
      & MRPP & -0.2407  & -0.2307  & -0.2875  & \textbf{-0.0039} & \textbf{-0.1113} & -0.0960  & \textbf{0.2392} \\
    \midrule
    \multicolumn{1}{c}{\multirow{4}[2]{*}{Semantic\newline{}}} & base & -0.1582  & -0.1620  & -0.2949  & -0.0687  & -0.1287  & -0.0897  & 0.3541  \\
      & CoT & \textbf{-0.1192} & \textbf{-0.1136} & \textbf{-0.2083} & \textbf{-0.0488} & \textbf{-0.0448} & \textbf{-0.0454} & \textbf{0.3494} \\
\bottomrule
    \end{tabular}%
  \label{tab:corr-cnn}%
\end{table*}%

%% file: tables/ece-llama.tex
\begin{table*}[htb]
  \centering
  \small
  \caption{ECE results of \texttt{Llama-3.1-8B-Instruct} across different datasets. The bolded values indicate cases with reduced calibration.}
    \begin{tabular}{clccccccc}
    \toprule
      &   & PPL & Sharpness & \makecell{Eigen\\Score} & \makecell{SelfCk\\-Prompt} & \makecell{SelfCk\\-NLI} & \makecell{Hidden\\Score} & \makecell{Attn\\Score} \\
    \midrule
    \multicolumn{1}{c}{\multirow{4}[1]{*}{\makecell{Textual \\ HaluEval}}} & base & 16.89  & 17.00  & 24.90  & 8.40  & 23.64  & 25.90  & 40.09  \\
      & CoT & 14.98  & 15.03  & 20.25  & \textbf{12.05} & 17.37  & \textbf{28.74} & 21.90  \\
      & LtM & 14.93  & 15.29  & \textbf{27.23} & \textbf{11.27} & 14.35  & \textbf{27.12} & 21.45  \\
      & MRPP & 14.95  & 15.31  & 24.70  & \textbf{11.10} & 15.64  & \textbf{28.64} & 18.63  \\
    \midrule
    \multicolumn{1}{c}{\multirow{4}[2]{*}{\makecell{Semantic \\ HaluEval}}} & base & 19.50  & 19.34  & 13.37  & 16.08  & 21.90  & 25.24  & 37.27  \\
      & CoT & \textbf{66.88} & \textbf{66.74} & \textbf{24.01} & \textbf{58.64} & \textbf{55.80} & \textbf{33.10} & \textbf{45.99} \\
      & LtM & 16.50  & 15.59  & \textbf{20.38} & \textbf{17.08} & 19.77  & 21.68  & 17.76  \\
\bottomrule
    \end{tabular}%
  \label{tab:ece-llama}%
\end{table*}%

%% file: tables/ece-mistral.tex
\begin{table*}[htb]
  \centering
  \small
  \caption{ECE results of \texttt{Mistral-7B-Instruct-v0.3} across different datasets. The bolded values indicate cases with reduced calibration.}
    \begin{tabular}{clccccccc}
      \toprule
      &   & PPL & Sharpness & \makecell{Eigen\\Score} & \makecell{SelfCk\\-Prompt} & \makecell{SelfCk\\-NLI} & \makecell{Hidden\\Score} & \makecell{Attn\\Score} \\
    \midrule
    \multicolumn{1}{c}{\multirow{4}[2]{*}{\makecell{Textual \\ HaluEval}}} & base & 12.80  & 12.75  & 23.43  & 19.43  & 17.78  & 18.71  & 23.51  \\
      & CoT & \textbf{17.06} & \textbf{17.10} & \textbf{26.82} & \textbf{38.75} & \textbf{19.96} & \textbf{18.93} & 14.34  \\
      & LtM & \textbf{18.26} & \textbf{18.24} & \textbf{29.60} & \textbf{40.78} & \textbf{18.42} & \textbf{20.62} & 14.58  \\
      & MRPP & \textbf{18.63} & \textbf{18.67} & \textbf{30.54} & \textbf{43.73} & \textbf{20.05} & \textbf{19.71} & 13.19  \\
    \midrule
    \multicolumn{1}{c}{\multirow{4}[2]{*}{\makecell{Semantic \\ HaluEval}}} & base & 11.26  & 11.23  & 19.60  & 26.37  & 23.74  & 14.24  & 18.15  \\
      & CoT & \textbf{56.92} & \textbf{56.91} & 17.20  & 22.25  & \textbf{61.50} & \textbf{47.64} & \textbf{56.51} \\
      & LtM & \textbf{60.89} & \textbf{60.89} & \textbf{25.75} & 21.92  & \textbf{67.90} & \textbf{46.80} & \textbf{56.03} \\
\bottomrule
    \end{tabular}%
  \label{tab:ece-mistral}%
\end{table*}%

%% file: tables/ece-r1.tex
% Table generated by Excel2LaTeX from sheet 'ece'
\begin{table*}[htbp]
  \centering
  \small
  \caption{ECE results of \texttt{DeepSeek-R1-Distill-Llama-8B} across different datasets. The bolded values indicate cases with reduced calibration.}
    \begin{tabular}{ccrrrrrrr}
    \toprule
      &   & \multicolumn{1}{c}{PPL} & \multicolumn{1}{c}{Sharpness} & \makecell{Eigen\\Score} & \makecell{SelfCk\\-Prompt} & \makecell{SelfCk\\-NLI} & \makecell{Verbalized\\Certainty} & \makecell{Hidden\\Score} \\
    \midrule
    \multicolumn{1}{c}{\multirow{4}[2]{*}{\makecell{Textual \\ HaluEval}}} & base & 0.2133  & 0.2098  & 0.4199  & 0.5159  & 0.3615  & 0.3567  & 0.4295  \\
      & CoT & 0.2088  & 0.2060  & \textbf{0.4256} & \textbf{0.6087} & 0.1700  & 0.3016  & \textbf{0.4706} \\
      & LtM & 0.1774  & 0.1789  & 0.4124  & \textbf{0.6148} & 0.1575  & 0.3186  & \textbf{0.4798} \\
      & MRPP & \textbf{0.2133} & 0.2053  & 0.4124  & \textbf{0.6052} & 0.1911  & 0.3225  & \textbf{0.4860} \\
    \midrule
    \multicolumn{1}{c}{\multirow{4}[2]{*}{\makecell{Semantic \\ HaluEval}}} & base & 0.2590  & 0.2607  & 0.1069  & 0.1943  & 0.2234  & 0.1299  & 0.1212  \\
      & CoT & \textbf{0.4416} & \textbf{0.4446} & \textbf{0.2364} & 0.0880  & \textbf{0.7213} & \textbf{0.2962} & \textbf{0.1452} \\
      & LtM & \textbf{0.4757} & \textbf{0.4789} & \textbf{0.2534} & 0.0805  & \textbf{0.7346} & \textbf{0.3158} & \textbf{0.1396} \\
\bottomrule
    \end{tabular}%
  \label{tab:ece-r1}%
\end{table*}%

%% file: tables/ece-70b.tex
% Table generated by Excel2LaTeX from sheet 'ece'
\begin{table*}[htbp]
  \centering
  \small
  \caption{ECE results of \texttt{Llama-3.1-70B-Instruct} across different datasets. The bolded values indicate cases with reduced calibration.}
    \begin{tabular}{ccrrrrrrr}
    \toprule
      &   & \multicolumn{1}{c}{PPL} & \multicolumn{1}{c}{Sharpness} & \makecell{Eigen\\Score} & \makecell{SelfCk\\-Prompt} & \makecell{SelfCk\\-NLI} & \makecell{Hidden\\Score} & \makecell{Attn\\Score} \\
    \midrule
    \multicolumn{1}{c}{\multirow{4}[2]{*}{\makecell{Textual \\ HaluEval}}} & base & 0.1653  & 0.1650  & 0.1446  & 0.0923  & 0.1761  & 0.3062  & 0.3536  \\
      & CoT & \textbf{0.2211} & \textbf{0.2218} & \textbf{0.3481} & \textbf{0.2309} & \textbf{0.2143} & \textbf{0.4463} & \textbf{0.3715} \\
      & LtM & \textbf{0.1930} & \textbf{0.1918} & \textbf{0.2600} & \textbf{0.2045} & 0.1752  & \textbf{0.4767} & \textbf{0.3642} \\
      & MRPP & \textbf{0.2225} & \textbf{0.2228} & \textbf{0.3641} & \textbf{0.2463} & \textbf{0.2109} & \textbf{0.4350} & \textbf{0.4757} \\
    \midrule
    \multicolumn{1}{c}{\multirow{4}[2]{*}{\makecell{Semantic \\ HaluEval}}} & base & 0.1075  & 0.0977  & 0.1295  & 0.2224  & 0.2170  & 0.1082  & 0.2587  \\
      & CoT & \textbf{0.1507} & \textbf{0.1493} & 0.1218  & 0.2018  & 0.2384  & 0.1038  & 0.1468  \\
      & LtM & \textbf{0.1828} & \textbf{0.1989} & \textbf{0.1495} & \textbf{0.2231} & 0.3455  & \textbf{0.1234} & 0.1492  \\
\bottomrule
    \end{tabular}%
  \label{tab:ece-70b}%
\end{table*}%

%% file: tables/ece-cnn.tex
% Table generated by Excel2LaTeX from sheet 'ece'
\begin{table*}[htbp]
  \centering
  \small
  \caption{ECE results of CNN/Daily Mail dataset across different LLMs. The bolded values indicate cases with reduced calibration.}
    \begin{tabular}{ccrrrrrrr}
    \toprule
      &   & \multicolumn{1}{c}{PPL} & \multicolumn{1}{c}{Sharpness} & \makecell{Eigen\\Score} & \makecell{SelfCk\\-Prompt} & \makecell{SelfCk\\-NLI} & \makecell{Hidden\\Score} & \makecell{Attn\\Score} \\
    \midrule
    \multicolumn{9}{c}{Llama-3.1-8B-Instruct} \\
    \multicolumn{1}{c}{\multirow{4}[1]{*}{Textual}} & base & 0.1529  & 0.1487  & 0.1974  & 0.1831  & 0.1882  & 0.1557  & 0.2329  \\
      & CoT & 0.1209  & 0.1227  & 0.1929  & \textbf{0.2008} & \textbf{0.2377} & 0.1553  & 0.1951  \\
      & LtM & 0.1321  & 0.1198  & 0.1484  & \textbf{0.1952} & \textbf{0.2605} & \textbf{0.1896} & 0.1880  \\
      & MRPP & 0.1237  & 0.1319  & \textbf{0.2409} & \textbf{0.1903} & \textbf{0.2169} & \textbf{0.1658} & 0.2272  \\
    \midrule
    \multicolumn{1}{c}{\multirow{4}[2]{*}{Semantic}} & base & 0.1805  & 0.1753  & 0.1629  & 0.1897  & 0.2506  & 0.1750  & 0.1284  \\
      & CoT & 0.1570  & 0.1568  & \textbf{0.3774} & 0.1668  & 0.2370  & \textbf{0.2014} & \textbf{0.2173} \\
\bottomrule
    \end{tabular}%
  \label{tab:ece-cnn}%
\end{table*}%

%% file: tables/tqa-roc-full.tex
\begin{table*}[htb]
  \centering
  \small
  \caption{AUROC results of TruthfulQA. The bolded values indicate cases with reduced performance.}
    \begin{tabular}{cccccccccc}
    \toprule
      &   & PPL & Sharpness & \makecell{Eigen\\Score} & \makecell{SelfCk\\-Prompt} & \makecell{SelfCk\\-NLI} & \makecell{Verbalized\\Certainty} & \makecell{Hidden\\Score} & \makecell{Attn\\Score} \\
    \midrule
      &  & \multicolumn{8}{c}{\textit{Llama-3.1-8B-Instruct}} \\
    \multirow{4}[2]{*}{Info.} & base & 65.37  & 65.74  & 62.61  & 55.57  & 56.28  & 61.70  & 57.74  & 52.13  \\
      & CoT & \textbf{54.87} & \textbf{54.61} & \textbf{54.08} & \textbf{50.22} & 56.55  & \textbf{53.64} & \textbf{50.82} & 57.46  \\
      & LtM & \textbf{53.69} & \textbf{53.49} & 66.49  & 56.37  & \textbf{55.95} & \textbf{57.03} & 61.21  & 60.56  \\
      & MRPP & \textbf{56.15} & \textbf{55.81} & \textbf{55.76} & 55.62  & 63.02  & 70.01  & \textbf{51.33} & 80.12  \\
    \midrule
    \multirow{4}[2]{*}{Truth.} & base & 50.35  & 50.36  & 54.34  & 59.17  & 60.31  & 54.69  & 52.43  & 50.78  \\
      & CoT & 53.06  & 53.61  & \textbf{51.67} & \textbf{54.79} & \textbf{58.23} & \textbf{53.23} & 54.84  & \textbf{50.14} \\
\bottomrule
    \end{tabular}%
  \label{tab:tqa-roc-full}%
\end{table*}%

%% file: tables/qwen-judge.tex
% Table generated by Excel2LaTeX from sheet 'qwen'
\begin{table*}[htbp]
  \centering
  \small
  \caption{To mitigate potential biases introduced by the LLaMA-based judge model, we also fine-tune the evaluation model using Qwen. The evaluation results based on the Qwen judge model are consistent with our previous findings, confirming the robustness of our conclusions.}
    \begin{tabular}{cccccccccc}
    \toprule
      &   & PPL & Sharpness & \makecell{Eigen\\Score} & \makecell{SelfCk\\-Prompt} & \makecell{SelfCk\\-NLI} & \makecell{Verbalized\\Certainty} & \makecell{Hidden\\Score} & \makecell{Attn\\Score} \\
    \midrule
          &  & \multicolumn{8}{c}{\textit{Llama-3.1-8B-Instruct}} \\
        \multirow{4}[2]{*}{Info.} & \multicolumn{1}{p{2.25em}}{base} & 56.02 & 56.52 & 61.56 & 54.35 & 65.94 & 67.57 & 72.12 & 71.36 \\
      & \multicolumn{1}{p{2.25em}}{CoT} & \textbf{54.04} & \textbf{55.31} & 72.77 & \textbf{53.61} & \textbf{61.02} & \textbf{66.32} & \textbf{51.55} & 74.78 \\
      & \multicolumn{1}{p{2.25em}}{LtM} & 59.23 & 58.75 & 65.92 & \textbf{50.21} & \textbf{57.74} & \textbf{59.63} & \textbf{64.98} & \textbf{58.63} \\
      & \multicolumn{1}{p{2.25em}}{MRPP} & 58.6 & 62.07 & 67.38 & 82.95 & \textbf{55.03} & \textbf{52.00} & \textbf{70.88} & \textbf{69.74} \\
    \midrule
    \multirow{4}[2]{*}{Truth.} & \multicolumn{1}{p{2.25em}}{base} & 56.31 & 57.33 & 67.49 & 56.06 & 53.19 & 52.64 & 75.18 & 77.3 \\
      & \multicolumn{1}{p{2.25em}}{CoT} & 56.76 & \textbf{55.54} & \textbf{64.28} & \textbf{52.61} & \textbf{50.32} & \textbf{52.31} & \textbf{60.81} & \textbf{63.94} \\
\bottomrule
    \end{tabular}%
  \label{tab:qwen-judge}%
\end{table*}%